\def\WMPAIarxiv{1}
\documentclass{article}

\ifdefined\WMPAIarxiv
  \usepackage[preprint]{neurips_2026}
\else
  \usepackage[final,sglblindworkshop]{neurips_2026}
  \workshoptitle{World Models in Physical AI}
\fi

\usepackage[utf8]{inputenc}
\usepackage[T1]{fontenc}
\usepackage{amsfonts}
\usepackage{amsmath}
\usepackage{amssymb}
\usepackage{array}
\usepackage{booktabs}
\usepackage{graphicx}
\usepackage{hyperref}
\usepackage{microtype}
\usepackage{nicefrac}
\usepackage{needspace}
\usepackage{placeins}
\usepackage{url}
\usepackage{xcolor}
\hypersetup{hidelinks,
  pdftitle={When Does Latent Prediction Error Support Physical Surprise Claims in Video World Models?},
  pdfauthor={Anton Hristov; Aleksandar Slavov; Vasil Vasilev}}
\graphicspath{{figures/}}

\title{When Does Latent Prediction Error Support Physical Surprise Claims in Video World Models?}
\author{%
Anton Hristov\textsuperscript{1}\thanks{All three authors contributed equally.} \quad
Aleksandar Slavov\textsuperscript{2}\footnotemark[1] \quad
Vasil Vasilev\textsuperscript{1}\footnotemark[1]\\
\textsuperscript{1}Technology School Electronic Systems (TUES)\\
\textsuperscript{2}Baba Tonka High School of Mathematics
}

\begin{document}
\maketitle

\begin{abstract}
\color{black}
Prediction errors from video world models are often treated as physical surprise, but they can also reflect appearance, static content, motion, or ordinary prediction difficulty. We present an evaluation framework that separates what frozen representations contain, how they organize appearance and physical differences, and what changes a model's own prediction error. On IntPhys2, a benchmark of physically possible and impossible events, we compare representations from latent prediction, masked pixel reconstruction, and simple visual statistics. Appearance remains easy to decode, while V-JEPA2 exposes modest information about whether events are possible or impossible. Looking at the latent geometry, appearance changes become smaller compared with differences between videos, their directions line up more consistently across scenes, and their overlap with physical changes decreases in deeper layers. Across action-conditioned models, incorrect actions raise prediction error, while appearance changes also produce substantial responses. In Physion++ video clips, repeating a single frame can produce higher error than the original clip, while changing color throughout the clip can lower it. Using Kubric, a simulation and rendering framework, we find that impossible events can produce lower error than possible events. A fixed error threshold also incorrectly flags many possible events as impossible. A model can therefore represent physical information and respond to incorrect actions without its native prediction error reliably measuring physical surprise.
\end{abstract}
\section{Introduction}
\label{sec:introduction}

Video world models are expected to capture regularities that make visual futures predictable. Prediction errors and related scores are consequently used to assess physical plausibility and identify unexpected events \citep{smith2019adept,piloto2022plato,riochet2018intphys,dai2023xvoe,garrido2025intuitive,bordes2025intphys2,maes2026leworldmodel}. Such scores also guide the selection of physically plausible generated videos \citep{baldassarre2025dinoworld,yuan2026wmreward}. Understanding what they measure matters when errors are used to locate events, compare videos, or reward trajectories.

Let $f_C$ and $f_T$ construct context and target latent representations, also called features, from observations $x$. Both functions include the model's encoding and token-selection steps. For a predictor $p$, optional actions $a$, and a scoring distance $d$, the prediction-error score is
\[
S(x,a)=d\!\left(p(f_C(x),a),\,f_T(x)\right).
\]
This score measures prediction difficulty in the chosen representation. A large value may reflect physical inconsistency, appearance, static content, motion, or ordinary model error. Physical information may also be accessible to a classifier even when the predictor's error responds primarily to other properties of the video.

We first test whether linear classifiers can distinguish physically possible from impossible events using frozen representations. We also test how well appearance information can be recovered from the same representations. IntPhys2 is a synthetic video benchmark pairing physically possible events with violations such as an object passing through a solid obstacle \citep{bordes2025intphys2}. We compare V-JEPA2-G, trained to predict latent representations, with VideoMAE2-G, trained to reconstruct masked pixels, and simple visual statistics. Linear classifiers readily recover appearance labels from all three. V-JEPA2-G provides a modest advantage when classifying events as possible or impossible.

We then examine how appearance and physical differences are organized across network depth. Changes in hue, brightness, and object color let us measure how far representations move, how consistently they move in the same direction across scenes, and how much those directions overlap with differences between possible and impossible events. In both V-JEPA2-G and VideoMAE2-G, appearance shifts become smaller relative to distances between unrelated videos. At deeper layers, the shifts line up more than those from randomly paired videos and overlap less with physical changes, though these trends sometimes reverse between layers. Appearance labels nevertheless remain easy to recover at the evaluated probe layers. We also test whether representation distances help predict changes in V-JEPA2-G's prediction error. In a regression analysis, they add little beyond scene conditions and simple visual statistics.

Finally, we test prediction error under controlled changes to actions, appearance, motion, and physical validity. Across three action-conditioned world models, incorrect actions raise error while the observed frames are held fixed. Color changes also alter error without changing the simulated physical state. In Physion++ videos of physical interactions, repeating a single frame can increase error, whereas changing color throughout a clip can lower it \citep{tung2023physionpp}. Using Kubric, a simulation and rendering framework, we test stopping and falling with and without support beneath an object \citep{greff2022kubric}. Mean error is higher for stopping in both settings, even though stopping is lawful only when support is present. A fixed error threshold also flags many lawful events as impossible. These results show why recoverable physical information and action sensitivity alone are insufficient to interpret prediction error as physical surprise.

\section{Related work}
\label{sec:related-work}

\paragraph{Predictive video representations.}
Joint-embedding predictive architectures learn by predicting latent representations rather than pixels \citep{lecun2022path,bardes2024feature}. V-JEPA2 extends this approach to video understanding, prediction, and planning \citep{assran2025vjepa2}, while VideoMAE and VideoMAE2 learn representations through masked pixel reconstruction \citep{tong2022videomae,wang2023videomaev2}. Action-conditioned models such as LeWorldModel, DINO-WM, and JEPA-WM predict visual representations from observations and actions \citep{maes2026leworldmodel,zhou2024dinowm,terver2026jepawm}. These objectives provide different representations and native error scales. Our comparisons retain each released pipeline and test the meaning of its error through controlled interventions.

\paragraph{Physical information and latent geometry.}
Inverse-dynamics probes recover actions from video representations \citep{yeom2026actionrelevant}; robotic evaluations distinguish representations useful for reconstruction from those useful for planning \citep{nilaksh2026reconstruction}. Layerwise probes and subspace analyses reveal physics-related structure in V-JEPA2 and VideoMAE2 \citep{joseph2026interpreting,punzo2026layerwise}. Geometry regularization can improve multistep prediction \citep{xia2026cloning}, and PhyLatent studies physical invariance and identifiability under appearance and dynamics changes \citep{zeng2026phylatent}. We connect this representation-level evidence to a separate question: whether measured geometry predicts native-error differences in the same model and setting.

\paragraph{Physical prediction and violation of expectation.}
IntPhys and IntPhys2 evaluate possible versus impossible events \citep{riochet2018intphys,bordes2025intphys2}; Physion and Physion++ evaluate physical-outcome prediction and inference of latent physical properties \citep{bear2021physion,tung2023physionpp}. Violation-of-expectation models use predicted-versus-observed discrepancies to identify unexpected physical events \citep{smith2019adept,piloto2022plato,dai2023xvoe,garrido2025intuitive}. Related evaluations address counterfactual consistency and physical plausibility in generated videos \citep{begiristain2026cronos,baldassarre2025dinoworld,yuan2026likephys,yuan2026wmreward}. We focus on the specificity and transfer of a fixed native score: whether physical violations increase error across controlled changes in motion, appearance, and context.

\Needspace{12\baselineskip}
\section{Evaluation framework}
\label{sec:framework}

We retain each evaluated pipeline's encoders, token selection, normalization, and native distance. Context-only prediction and encoding a full clip before selecting context tokens are distinct computations (Section~\ref{sec:results-visible-events}). Native score magnitudes are compared only within a model.

Table~\ref{tab:audit-gates} separates evidence about representations from evidence about the score. Readout and geometry characterize encoded information; the remaining checks test predictive sensitivity and physical decisions. For threshold-based detection, a lawful clip above the threshold is a false alarm, and an impossible clip below it is a missed violation.

\begin{table}[htbp]
\centering
\small
\setlength{\tabcolsep}{4pt}
\renewcommand{\arraystretch}{1.08}
\caption{Six complementary checks for interpreting prediction error as physical surprise. Calibration here means transfer of a fixed decision threshold, rather than calibration of predicted probabilities.}
\label{tab:audit-gates}
\begin{tabular}{@{}>{\raggedright\arraybackslash}p{0.17\linewidth}>{\raggedright\arraybackslash}p{0.77\linewidth}@{}}
\toprule
Check & Controlled comparison and measured property \\
\midrule
Readout & Train a linear classifier on frozen representations and test it on scenes excluded from training. Measure how accurately it distinguishes physically possible from impossible events. \\
\addlinespace[4pt]
Geometry & Measure appearance and physical shifts across layers. Relative size shows how strongly appearance changes affect representations compared with differences between unrelated videos. Stronger alignment than in shuffled pairs indicates a consistent response across scenes. Greater overlap than with random subspaces of the same dimension indicates shared directions between appearance and physical changes. \\
\addlinespace[4pt]
Action & Keep the observed outcome fixed and compare predictions made with correct and mismatched actions. Check whether incorrect actions increase prediction error. \\
\addlinespace[4pt]
Appearance/static & Change appearance without changing the simulated physical state, or repeat a single frame throughout a clip. Compare prediction error with the original clip. \\
\addlinespace[4pt]
Physical validity & Compare possible and impossible events while varying motion or appearance. Check whether impossible events consistently receive higher error within each condition. \\
\addlinespace[4pt]
Calibration & Choose an error threshold on separate calibration data and apply it unchanged to new scenes and appearances. Measure the rates of false alarms and missed violations. \\
\bottomrule
\end{tabular}
\end{table}

We evaluate V-JEPA2-G on controlled Kubric videos and IntPhys2, V-JEPA2-L on Physion++ Mass, and released action-conditioned models on TwoRoom and PushT. The IntPhys2 comparison includes VideoMAE2-G and low-level visual statistics. DINO-WM and JEPA-WM share a frozen DINOv2 encoder \citep{oquab2024dinov2}, so their comparison varies only the predictor.

Uncertainty is estimated by bootstrap over independent configurations, Physion++ matched pairs within scenarios, IntPhys2 scenes, or action-test episodes. All conditions belonging to one unit remain together. Bonferroni-adjusted intervals are used for the specified families of contrasts. Thresholds and regression coefficients are fitted on data separate from evaluation. Full protocols, fitting and split histories, and reproducibility records are in the appendices. Code and numeric experiment records are available at \url{https://github.com/3-ml/jepa-surprise} (Appendix~\ref{app:resources-provenance}).

\FloatBarrier
\section{Results}

\subsection{Physical-violation ordering and threshold transfer}
\label{sec:results-simulator-controls}

To separate physical validity from trajectory identity, we render a falling sphere above either a solid platform or an opening. Each setting has two continuations: the sphere stops at platform height or falls below it. Stopping is lawful on the solid platform and unsupported over the opening; falling is lawful through the opening and violates solidity when it passes through the platform. Figure~\ref{fig:support-examples} in the appendix shows the rendered construction. The 256 configurations cross these four cases with the original appearance, a $30^\circ$ hue shift throughout the clip (persistent), or the same shift starting at the target interval (abrupt), giving 3,072 clips. Stopping trajectories match across settings, as do falling trajectories. Within each setting and appearance, the two continuations share context frames and predictor outputs. Support appearance, shadows, and occlusion differ across settings. The predictor observes the first 12 of 48 frames; native error is mean absolute difference from normalized EMA-target tokens for the remaining 36 frames.

\begin{figure}[htbp]
\centering
\includegraphics[width=\linewidth]{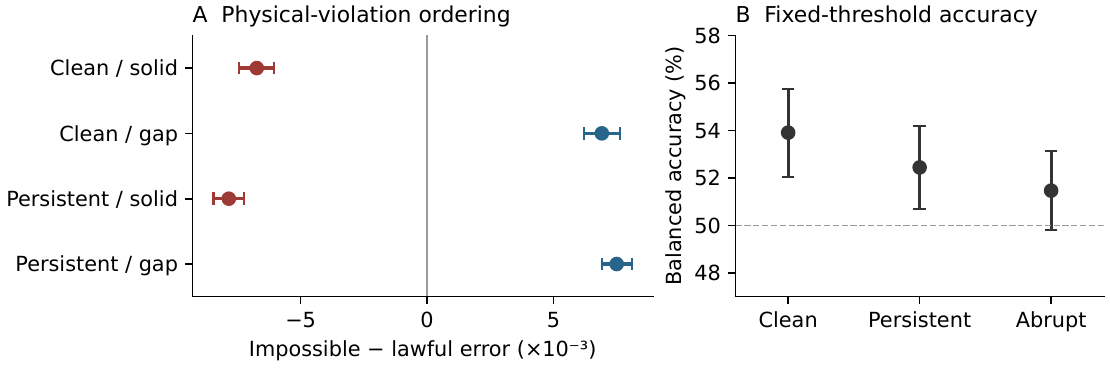}
\caption{V-JEPA2-G on 256 configurations. Left: impossible minus possible native error, with 98.75\% configuration-bootstrap intervals adjusted for four means. The sign reverses between solid support and a gap. Right: balanced accuracy under a single threshold fitted on separate calibration scenes, with descriptive 95\% configuration-bootstrap intervals. The dashed line marks 50\%.}
\label{fig:final256}
\end{figure}

V-JEPA2-G gives stopping higher mean error in both settings. Consequently, impossible-minus-possible error is \emph{negative} on solid support and \emph{positive} over the gap (Figure~\ref{fig:final256}). The native score's ordering follows the trajectory across this reversal of physical validity (Appendix Figure~\ref{fig:appendix-final256-crossover}).

The reversal holds under both clean and persistent-hue appearance; all four adjusted intervals exclude zero in their respective directions. Averaging the two settings would conceal the opposite orderings. Appendix~\ref{app:final256-estimands} reports the complete margins and intervals.

A single threshold fitted on 24 separate clean calibration configurations yields 53.9\% balanced accuracy [52.1, 55.8] on clean evaluation clips, with equal weight to both support settings. Performance differs sharply by setting: 80.7\% over gaps versus 27.1\% on solid support. The threshold falsely flags about 69\% of lawful clips, including every lawful solid-support continuation (Appendix Figure~\ref{fig:appendix-final256-distributions}). Thus the above-chance average coexists with a large false-alarm rate. Figure~\ref{fig:final256} also shows transfer under persistent and abrupt hue; Appendix~\ref{app:final256-estimands} gives all per-setting rates and the threshold definition.

\subsection{Static-content and persistent-appearance controls}
\label{sec:results-visible-events}

We evaluate V-JEPA2-L's full-clip masked-prediction score on the 192 official Physion++ Mass videos. The encoder processes each 16-frame window before context tokens are selected, so target frames can influence the predictor. The static diagnostic compares every complete, nonoverlapping window with versions repeating its first, middle, or last frame. These still-frame controls have higher mean error than the original videos. Across videos, the mean share of windows scoring no higher than their average static baseline is approximately 84\%. The mean original-minus-static difference is negative in every scenario, and the layout-cluster sensitivity check agrees. Uncertainty is based on scenario-stratified matched pairs, with windows averaged within videos, not treated as independent observations. Repeating frames changes both content and history, so this comparison neither isolates motion's causal effect nor supplies a universal score correction.

Persistent appearance changes can also \emph{lower} error. This separate experiment uses one fixed middle 16-frame window per video. Hue shifts and pixel darkening applied throughout the window both lower mean error relative to the original and to their target-only counterparts (Table~\ref{tab:physion-appearance}). Changes restricted to the final four frames have uncertain original-relative effects: their adjusted intervals include zero. The persistent-versus-target-only hue ordering reverses in Collision, so the overall mean does not describe every scenario.

The normalized L1 score averages spatial-token errors over the two target frame-pairs (temporal tubelets). Because the encoder sees the full window, target-only edits can affect the predictor despite unchanged context pixels. These are diagnostics of the evaluated masked-prediction implementation. Appendix~\ref{app:physion-full} gives the complete counts, intervals, and scoring details.

\begin{table}[htbp]
\centering
\small
\caption{V-JEPA2-L appearance effects on 96 matched Physion++ pairs, in native error units. Each entry subtracts the second condition from the first: negative means lower error, not a negative error score. Brackets are 99.17\% pair-bootstrap intervals, Bonferroni-adjusted for six comparisons. Persistent changes affect the whole clip; target-only changes affect its final four frames.}
\label{tab:physion-appearance}
\begin{tabular}{@{}lrr@{}}
\toprule
Contrast & Hue & Pixel darkening \\
\midrule
Persistent $-$ original & $-0.00312$ [$-0.00385$, $-0.00236$] & $-0.00270$ [$-0.00337$, $-0.00203$] \\
Target-only $-$ original & $-0.00069$ [$-0.00168$, $+0.00021$] & $+0.00059$ [$-0.00038$, $+0.00158$] \\
Persistent $-$ target-only & $-0.00243$ [$-0.00337$, $-0.00134$] & $-0.00329$ [$-0.00415$, $-0.00242$] \\
\bottomrule
\end{tabular}
\end{table}

\subsection{Physical-label and appearance readouts}
\label{sec:results-physical-structure}

Linear probes readily recover appearance labels. For V-JEPA2-G, physical-label accuracy under these probes is only modestly above chance: approximately 58\% for possible versus impossible events, against 50\% chance (Table~\ref{tab:intphys-readouts}). Its four-class appearance accuracy is approximately 98\%, but the two tasks do not provide a common scale for the amount of information encoded. The physics result shows that label information is accessible. Whether native prediction error detects violations is a separate question, which the later experiments test.

IntPhys2 tests object permanence, continuous motion, solidity, and constant object properties \citep{bordes2025intphys2}. For example, a solidity pair contrasts a solid object stopping at an obstacle with one passing through it. We compare V-JEPA2-G and VideoMAE2-G with PixelSummary (RGB statistics, color histograms, and frame differences). Five-fold scene cross-fitting keeps each scene's pairs and appearances together, with class-balanced linear probes and training-only feature scaling. Appendix~\ref{app:intphys-repro} documents layer selection and the post-hoc split correction.

\begin{table}[htbp]
\centering
\small
\caption{IntPhys2 classifiers evaluated on held-out scenes. Top rows show mean scene accuracy (percent correct); bottom rows subtract the comparator's accuracy from V-JEPA2-G's (percentage points). Physics uses all four appearances or clean clips only. Brackets are 95\% global scene-bootstrap intervals with classifier predictions held fixed.}
\label{tab:intphys-readouts}
\begin{tabular}{>{\color{black}}l>{\color{black}}c>{\color{black}}c>{\color{black}}c}
\toprule
Feature family or contrast & Appearance classification & Physics classification & Clean-only physics \\
\midrule
V-JEPA2-G & 97.7\% [95.7, 99.3] & 57.9\% [54.2, 61.8] & 54.8\% [51.3, 58.8] \\
VideoMAE2-G & 95.3\% [92.8, 97.6] & 49.2\% [47.3, 50.9] & 49.6\% [46.5, 52.2] \\
PixelSummary & 79.1\% [73.7, 84.5] & 50.0\% [48.9, 51.0] & 51.8\% [48.7, 54.8] \\
\midrule
V-JEPA2-G $-$ VideoMAE2-G & 2.4 pp [$-0.1$, 5.0] & 8.7 pp [4.4, 13.3] & 5.3 pp [0.4, 10.1] \\
V-JEPA2-G $-$ PixelSummary & 18.6 pp [13.5, 24.1] & 7.9 pp [4.1, 11.8] & 3.1 pp [$-1.3$, 8.3] \\
\bottomrule
\end{tabular}
\end{table}

For the all-appearance physics task, V-JEPA2-G's paired advantage over each comparator has an interval above zero. The clean-only comparison with PixelSummary remains uncertain. Table~\ref{tab:intphys-readouts} and Appendix~\ref{app:readout-figure} give the full comparisons. Architecture and training differ across models, so the comparison covers the released representations as a whole and does not isolate the JEPA objective.

\FloatBarrier
\subsection{Layerwise representation geometry}
\label{sec:results-layerwise-geometry}

\begin{figure}[htbp]
\centering
\includegraphics[width=\linewidth]{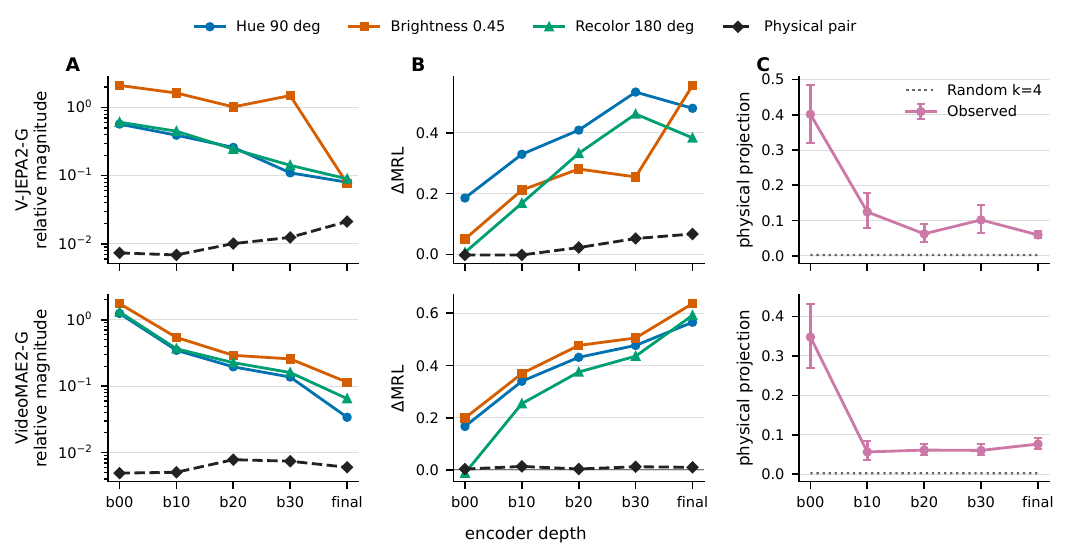}
\caption{Representation geometry on 57 IntPhys2 scenes. A: displacement relative to the median unrelated-video distance (log axis). B: directional coherence (mean resultant length, MRL) above shuffled correspondences. C: fraction of squared physical-pair displacement captured by a four-dimensional appearance subspace fitted on other scenes, with scene-bootstrap intervals and an equal-rank random reference.}
\label{fig:layerwise-structure}
\end{figure}

Both models show smaller final-layer appearance shifts \emph{relative to between-video differences} than at the first layer (Figure~\ref{fig:layerwise-structure}A). Relative magnitude divides each shift by the median unrelated-video distance at that layer. V-JEPA2-G's hue ratio falls from approximately 57\% to 8\%. The measures compare V-JEPA2-G and VideoMAE2-G on the same 57 IntPhys2 scenes.

Directional coherence measures alignment across videos. At later depths, appearance shifts are more aligned relative to shuffled source-transform pairings, whereas physical-change directions remain closer to that reference (Figure~\ref{fig:layerwise-structure}B).

Subspace overlap measures the fraction of squared physical-feature change projected into an appearance subspace fitted on other scenes. Final overlap is lower than initial overlap, but the trajectory is not monotonic. V-JEPA2-G's final overlap is approximately 6\%, above the roughly 0.28\% dimension-matched random reference (Figure~\ref{fig:layerwise-structure}C).

Appendix~\ref{app:intphys-estimand-geometry} gives the complete geometry and source-video retrieval results.

\subsection{Predicting native-error differences from geometry}
\label{sec:geometry-transfer}

We next test whether geometric measurements predict native-error differences in the crossed-support setting. Ordinary least squares (OLS) models fitted on 24 separate configurations predict physical margins (impossible minus lawful error) and appearance-induced score changes on the 256-configuration evaluation population. The baseline uses condition labels and static, motion, and shuffled-frame statistics; the expanded model adds relative representation distances and interactions with support or appearance timing. This compares geometry with information already available from the experimental conditions and visible changes.

Adding final-encoder geometry changes physical-margin $R^2$ from approximately 0.771 to 0.774, but the adjusted improvement interval includes both gains and losses. Some earlier locations worsen prediction. Here $R^2$ uses the error of predicting the fitting-set mean as its denominator, rather than the evaluation-set mean. For appearance-score differences, both models perform worse than that reference. Across all five locations and both tasks, no improvement interval lies entirely above zero (Appendix Tables~\ref{tab:geometry-physical}--\ref{tab:geometry-appearance}).

This post-hoc analysis gives limited evidence of added predictive value from the tested geometry features and linear models. Native prediction error is the world model's discrepancy between predicted and observed representations; the regression error here measures how well a separate model predicts differences in that discrepancy. Configuration-bootstrap intervals condition on the fitted regressions. Appendix~\ref{app:geometry-transfer} (Figure~\ref{fig:geometry-transfer}) gives the full protocol and analysis history.

\FloatBarrier
\subsection{Action and appearance sensitivity}
\label{sec:results-action-conditioning}

All three action-conditioned models use the supplied action: incorrect actions increase prediction error when the observed outcome is held fixed. We test 256 episodes per task or scorer under correct, zero, and reversed actions. LeWorldModel (LeWM) uses task-specific released checkpoints; DINO-WM and JEPA-WM reuse the PushT target frames and action interventions with their own preprocessing and predictors. Table~\ref{tab:native-error-summary} reports positive reversed-minus-correct contrasts for every model.

\begin{figure}[htbp]
\centering
\includegraphics[width=0.85\linewidth]{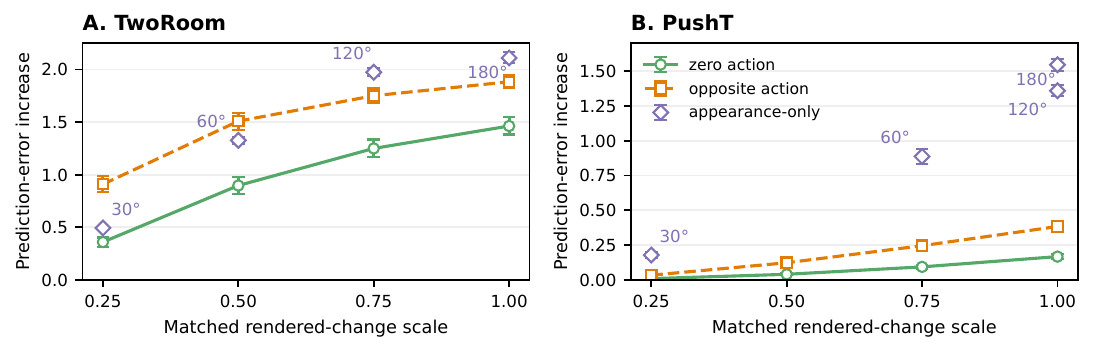}
\caption{LeWorldModel error changes. Action points compare incorrect and correct actions for the same outcome. Hue points compare changed color with the unchanged static target. The correct-action difference is zero and omitted.}
\label{fig:lewm-controls}
\end{figure}

Color changes also raise error while simulator state is held fixed. Figure~\ref{fig:lewm-controls} measures each hue change relative to the unchanged static target, and each action mismatch relative to the lawful prediction of the same moving target. Both interventions affect LeWM, with larger hue-induced increases at the two higher rendered-effect matches.

\begin{table}[htbp]
\centering
\small
\setlength{\tabcolsep}{8pt}
\caption{Same-target action effects: opposite minus lawful action at full strength, over 256 episodes per task or scorer. Brackets are 95\% episode-bootstrap intervals. Each model uses its native latent MSE; magnitudes are not comparable across models.}
\label{tab:native-error-summary}
\begin{tabular}{@{}lr@{}}
\toprule
Model/task & Opposite $-$ lawful \\
\midrule
LeWM TwoRoom & 1.884 [1.821, 1.946] \\
LeWM PushT & 0.3834 [0.3456, 0.4242] \\
DINO-WM PushT & 0.00364 [0.00304, 0.00428] \\
JEPA-WM PushT & 0.00351 [0.00293, 0.00411] \\
\bottomrule
\end{tabular}
\end{table}

The DINO-based models likewise have larger mean hue-induced than action-mismatch increases at the $60^\circ$/0.75 PushT match: 0.04977 versus 0.00307 for DINO-WM, and 0.01738 versus 0.00284 for JEPA-WM, each relative to its respective lawful baseline. These are descriptive means. Comparing absolute hue and opposite-action scores instead gives the reverse ordering because the moving lawful targets already have higher error. Appendix~\ref{app:pusht-baselines} (Figure~\ref{fig:pusht-baseline-waterfall}) decomposes that baseline difference and retains the absolute contrasts with their original intervals. These models therefore use the action and also respond to color. Matching the rendered RGB change of a hue edit to that of an action change does not make the two equally predictable to the model.

\FloatBarrier
\subsection{Task-associated color controls}
\label{sec:results-dynamics-specificity}

In a separate PushT study with simulator state fixed and RGB change matched, LeWM's error still depends on color choice. Agent-blue blocks score higher than novel-orange blocks; block-gray agents likewise exceed matched purple. The target-only blue/orange gap is positive in 233 of 256 episodes. Persistent changes reduce the mean gaps but retain both orderings (Appendix~\ref{app:pusht-role-color}).

For target-only edits, the prediction is unchanged and differences in target-encoder shifts account for almost all of the score gaps algebraically. Role colors can carry task information. These controls therefore show that the score depends on color. They leave open whether that color information helps predict physical outcomes and whether LeWorldModel encodes task roles.

\section{Discussion}
\label{sec:discussion}

Physical labels can be read out of V-JEPA2-G representations (about 58\% accuracy against 50\% chance), yet its native prediction error does not consistently track physical validity in the crossed-support test. Appearance remains decodable even as its relative displacement decreases across depth, while physical labels are modestly accessible under the evaluated probes. These findings describe the representation, but do not determine which video changes produce large native errors. The same-model regression makes that connection explicit: the selected geometry features add little predictive value beyond visual statistics and experimental conditions.

The native-error controls identify several sources of discrepancy. A score can remain high when motion is removed, or decrease after a persistent color change. In the crossed-support test, a stable trajectory ordering produces opposite physical-violation orderings and poor shared-threshold performance. Abrupt recoloring asks a different question: an unexpected color transition may itself violate object-property constancy. Persistent transformations remove that transition, allowing sensitivity to a stable appearance change to be examined separately. In PushT, the moving lawful targets already have higher error than the static ones. Comparing absolute scores can therefore rank the action change above the hue change, even though the hue increase over its own baseline is larger.

Table~\ref{tab:gate-verdicts} summarizes what each setting establishes. The evaluation framework turns the broad question of physical understanding into testable claims about a particular score. For a physical-surprise application, the decisive checks are violation ordering within relevant conditions and false-alarm and miss rates when a fixed decision rule transfers across them. These measurements connect the interpretation of prediction error to the decisions it is intended to support.
\begin{table}[htbp]
\centering
\footnotesize
\setlength{\tabcolsep}{3.8pt}
\renewcommand{\arraystretch}{1.0}
\caption{Evidence by model and population. S denotes support, I insufficient evidence, D appearance/static dependence, and -- untested. Contradicted marks lower mean error for impossible events than possible events on solid support. Appendix~\ref{app:gate-criteria} gives the criteria.}
\label{tab:gate-verdicts}
\begin{tabular}{@{}lcccccc@{}}
\toprule
Model / population & Readout & Geometry & Action & Appearance/static & Physical validity & Calibration \\
\midrule
V-JEPA2-G / IntPhys2 & S & S & -- & -- & -- & -- \\
VideoMAE2-G / IntPhys2 & I & S & -- & -- & -- & -- \\
V-JEPA2-G / crossed support & -- & -- & -- & D & Contradicted & I \\
V-JEPA2-L / Physion++ Mass & -- & -- & -- & D & -- & -- \\
LeWM / TwoRoom & -- & -- & S & D & I & -- \\
LeWM / PushT & -- & -- & S & D & I & -- \\
DINO-WM / PushT & -- & -- & S & D & I & -- \\
JEPA-WM / PushT & -- & -- & S & D & I & -- \\
\bottomrule
\end{tabular}
\end{table}

\section{Limitations}
\label{sec:limitations}

The controlled support experiment samples one procedural family with restricted cameras, colors, and lighting. It tests transfer across those conditions only; real-world and material generalization are untested. Material appearance can convey mass or friction cues, and greater training diversity may change appearance sensitivity. Those questions require additional data and controlled model comparisons.

The IntPhys2 readout and geometry-to-error transfer are post-hoc analyses. Their intervals condition on fitted models and omit refitting uncertainty; the readout also conditions on fold assignments. The tested linear geometry models do not establish a mechanism or the effect of removing appearance directions. Physion++ provides contact labels and no physical-violation labels, and its full-clip scoring implementation permits target information to affect context tokens. These distinctions constrain the interpretation of each result.

Model comparisons also vary architecture, pretraining, capacity, and native score. We do not estimate a general limit on planning or anomaly-detection performance. Application-specific acceptable error rates were not specified. Detailed split histories, failed compatibility checks, and excluded follow-ups are retained in the appendix.

\paragraph{\normalfont Broader impact.}
Prediction errors can serve as measures of physical surprise when controlled tests establish what they respond to. Their use as rewards or safety signals requires validation in the intended task and environment.
\label{end-main-text}
\clearpage
\bibliographystyle{plainnat}
\bibliography{references}

\clearpage
\appendix
% Insert immediately after \appendix and before \end{document}.
\section{Evidence map and interpretation}

\label{app:evidence-interpretation}
The experiments deliberately answer different questions. Table~\ref{tab:evidence-map} records the independent unit, evidential status, supported inference, and stopping point for each block. The main conclusion depends on their separation: representation accessibility, representation geometry, action-conditioned sensitivity, and error specificity are not interchangeable.

\begin{table}[htbp]
\centering
\small
\setlength{\tabcolsep}{3pt}
\caption{Evidence map including the completed crossed-support and Physion++ follow-ups. Status labels bound the valid inference; the older pilots remain historical.}
\label{tab:evidence-map}
\begin{tabular}{@{}>{\raggedright\arraybackslash}p{0.17\linewidth}>{\raggedright\arraybackslash}p{0.14\linewidth}>{\raggedright\arraybackslash}p{0.28\linewidth}>{\raggedright\arraybackslash}p{0.31\linewidth}@{}}
\toprule
Evidence block & Unit/status & Supports & Does not support \\
\midrule
{\color{black}IntPhys2 readout} & {\color{black}57 scenes / 114 pairs; post-hoc five-fold scene cross-fit} & {\color{black}Modest out-of-fold physics-label accessibility advantage under the evaluated probe budget} & {\color{black}Preregistered confirmation, native-error specificity, calibrated surprise, or general physical reasoning} \\
{\color{black}IntPhys2 frozen geometry} & {\color{black}57 scenes / 114 pairs; analysis fixed on development scenes, then run once on the sealed test set} & {\color{black}Controlled relative magnitude, null-adjusted direction, and cross-fitted subspace overlap for the evaluated perturbations/models} & {\color{black}Semantic disentanglement, causal factorization, residual specificity, calibration, or general physical understanding} \\
{\color{black}V-JEPA2-G multi-dose geometry} & {\color{black}38 development scenes / 76 pairs; exploratory, run after the sealed test} & {\color{black}How each transformation moves the features, and whether a local appearance basis captures more physical displacement than an endpoint basis of the same size} & {\color{black}Confirmation, hue circularity, semantic axes, or generality beyond tested factors} \\
LeWM / DINO-WM action controls & 256 episodes per task or scorer; confirmatory & Action-conditioned prediction sensitivity in LeWM TwoRoom, LeWM PushT, and DINO-WM PushT & Specificity of large error to dynamics or physical violation \\
{\leavevmode\color{black}JEPA-WM PushT action control} & {\leavevmode\color{black}256 episodes; post-hoc exploratory} & {\leavevmode\color{black}Action sensitivity and qualitative robustness of the DINO-WM ordering under a revised predictor recipe over the same encoder family} & {\leavevmode\color{black}Independent encoder-family replication, causal attribution to one model choice, broad cross-model generality, or calibrated physical surprise} \\
LeWorldModel appearance controls & 256 fresh PushT episodes; confirmatory plus exact decomposition & Appearance sensitivity, color-direction anisotropy, and partial persistent transport & Pure nuisance invariance or role-aware dynamics \\
Kubric crossed support & 256 configurations; frozen final test & Opposite native-error ordering across solid/gap worlds; fixed-threshold performance & Absence of physical knowledge, general utility ceiling, or real-world transfer \\
Geometry transfer & 24 development / 256 final; post-hoc & Limited incremental prediction, including negative gains & Causal mechanism, appearance-removal benefit, or fresh confirmation \\
Physion++ full population & 96 pairs / 192 videos; inspected diagnostic & Static and appearance dependence of the specified full-clip score & Violation accuracy, causal mediation, or context-only forecasting \\
Historical pilots & Kubric8; Physion10 / 273 windows & Exploratory displacement and static-content diagnostics & Equivalence, population false positives, or generality \\
\bottomrule
\end{tabular}
\end{table}
\FloatBarrier

\paragraph{Criteria for the main-text evidence table.}
\label{app:gate-criteria}
In Table~\ref{tab:gate-verdicts}, support refers to the inference named in Table~\ref{tab:audit-gates}. IntPhys2 readout support is post-hoc, and geometry support concerns measured representation structure. The separate crossed-support geometry-to-error regression is not a second test of that structural criterion. Its result is discussed in Section~\ref{sec:geometry-transfer}. The physical-validity check fails in the crossed-support experiment because violation margins have opposite signs across support conditions. Calibration remains insufficient: the fixed threshold makes frequent errors and no application-specific acceptable-error margin was specified. D records measured score dependence under appearance or static controls, without treating every such response as a failed physical judgment. The Physion++ entry concerns the full-clip score; the action-model entries use the historical controls and the baseline decomposition in Appendix~\ref{app:pusht-baselines}. Action support comes from same-target action contrasts, with JEPA-WM marked post-hoc. The excluded later two-model PushT continuation contributes no verdict.

\section{Representation readouts and geometry}
\subsection{IntPhys2 readout protocol and uncertainty}
\label{app:intphys-repro}
\label{app:intphys-readouts-details}
The corrected IntPhys2 sealed test contains 114 matched possible/impossible pairs from 57 held-out scenes, 228 source videos, and 912 appearance-conditioned samples. Each sample uses exactly 64 decoded source frames sampled by rounded linear spacing. The four appearance conditions are clean, 90-degree hue rotation, brightness multiplier 0.45, and 180-degree selective recoloring. V-JEPA2-G and VideoMAE2-G are extracted at five frozen depths (\texttt{block\_00}, \texttt{block\_10}, \texttt{block\_20}, \texttt{block\_30}, \texttt{final}); PixelSummary is a 100-dimensional low-level baseline of RGB summary, color-histogram, and frame-difference statistics. The probe is \texttt{StandardScaler} plus balanced logistic regression with $C=1.0$ and at most 2000 iterations. The post-hoc correction reported here sorts scenes by hash (seed 0) and splits them into five folds. The fold index (0 to 4) is also the classifier's random state. Fold $k$ is test, fold $(k+1)\bmod5$ is validation, and the remaining folds are training. Sealed test folds contain 12, 12, 11, 11, and 11 scenes; development test folds contain 8, 8, 8, 7, and 7, with all pair IDs and appearance variants from one scene held together. Primary layers preserve the original development choices: V-JEPA2-G \texttt{block\_20} for appearance and \texttt{block\_30} for physics, VideoMAE2-G \texttt{block\_30} for appearance, \texttt{block\_00} for all-appearance physics, and \texttt{block\_30} for clean-only physics, with PixelSummary using its only layer. The cross-fit development check retains V-JEPA2-G \texttt{block\_20} for appearance and \texttt{block\_30} for \texttt{physics\_all}; its all-appearance comparator deltas include zero. Table~\ref{tab:intphys-pipelines} summarizes the model-specific input and temporal pipelines.

\begin{table}[htbp]
\centering
\small
\setlength{\tabcolsep}{3pt}
\caption{Checkpoint-family feature pipelines. The comparison does not isolate training objective or architecture.}
\label{tab:intphys-pipelines}
\begin{tabular}{@{}lccccp{0.25\linewidth}@{}}
\toprule
Family & Input & Native window & Patch/tubelet & Layers & Temporal scope \\
\midrule
V-JEPA2-G & $64\times256^2$ & 64 frames & $16/2$ & 0,10,20,30,40 & one joint 64-frame clip \\
VideoMAE2-G & $4\times16\times224^2$ & 16 frames & $14/2$ & 0,10,20,30,40 & four independent chunks \\
PixelSummary & 64 frames & n/a & n/a & one & 100 RGB/histogram/difference statistics \\
\bottomrule
\end{tabular}
\end{table}

Both neural pipelines receive the same decoded frames, use ImageNet RGB normalization, and expose 1408-dimensional pooled features at matched relative depths. V-JEPA2-G resizes without cropping; VideoMAE2-G uses its pinned short-side resize and center crop. The shared coarse token grid is $4\times4\times4$, obtained by mean pooling native spatiotemporal tokens. 

For physics, correctness is averaged over samples within each out-of-fold pair and then over both pairs in each unique scene. Each of 10,000 bootstrap draws samples all 57 sealed scene values globally with replacement. Model deltas preserve fold, scene, and pair alignment before scene aggregation and paired resampling. These intervals reflect resampling of scenes with the trained probes held fixed. They do not include variation from refitting the probes or from a different fold assignment. Near- or below-chance comparator estimates are not interpreted as meaningful anti-physics signals.

\subsection{Linear-probe results}
\label{app:readout-figure}
\begin{figure}[htbp]
\centering
\includegraphics[width=\linewidth]{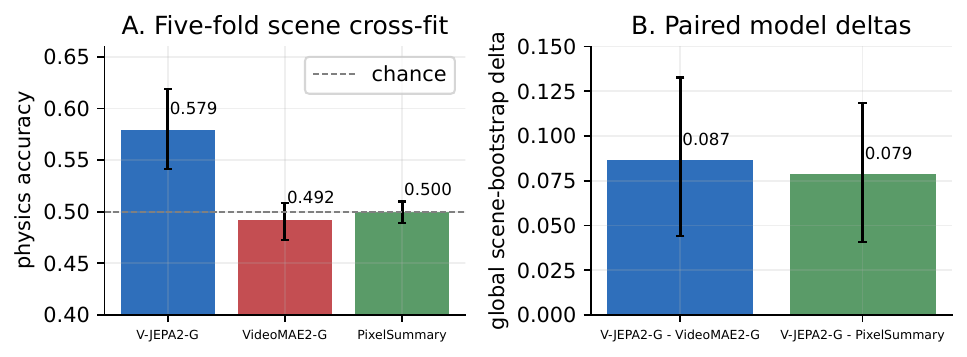}
\caption{Post-hoc five-fold IntPhys2 scene cross-fit on 57 scenes. Each scene is held out once; intervals and paired differences bootstrap the 57 scene aggregates while conditioning on fitted probes. They omit probe-refitting and fold-assignment uncertainty.}
\label{fig:intphys-crossfit}
\end{figure}

The main text reports all three classification tasks and paired model differences in Table~\ref{tab:intphys-readouts}. Figure~\ref{fig:intphys-crossfit} visualizes the same physics-readout estimates; it is not an additional experiment or an independent confirmation.
\FloatBarrier
\subsection{Geometry estimands and complete results}
\label{app:intphys-estimand-geometry}
{\color{black}\paragraph{Geometry estimands and nulls.} Let $\hat z_i^{(\ell)}=z_i^{(\ell)}/\lVert z_i^{(\ell)}\rVert_2$ and $d(x,y)=1-\cos(x,y)$. The unrelated clean scale and relative displacement magnitude are
\[
G_\ell=\mathop{\mathrm{median}}_{i,j:\,s_i\ne s_j}d(\hat z_i^{\mathrm{clean}},\hat z_j^{\mathrm{clean}}),
\qquad R_\ell(T;i)=\frac{d(\hat z_i^{\mathrm{clean}},\hat z_i^T)}{G_\ell},
\]
with the same $G_\ell$ used for matched physical pairs. For $\Delta_i^T=\hat z_i^T-\hat z_i^{\mathrm{clean}}$ and non-negligible $u_i^T=\Delta_i^T/\lVert\Delta_i^T\rVert_2$, directional coherence is the standard mean resultant length \citep{mardia1999directional},
\[
\mathrm{MRL}_\ell(T)=\left\lVert N^{-1}\sum_i u_i^T\right\rVert_2,
\qquad
\Delta\mathrm{MRL}=\mathrm{MRL}_{\mathrm{obs}}-\mathbb E[\mathrm{MRL}_{\mathrm{shuffle}}].
\]
Our operational control shuffles source-transform correspondence only across scenes, retaining the factor, model, and layer. An orthonormal appearance basis $U_{\mathrm{app},k}^{(\ell)}$ is obtained by standard SVD/PCA \citep{jolliffe2016pca} on training-scene displacements. For held-out $\Delta_i^{\mathrm{phys}}=\hat z_i^{\mathrm{impossible}}-\hat z_i^{\mathrm{possible}}$,
\[
P_{\mathrm{app},k}^{(\ell)}(i)=
\frac{\lVert (U_{\mathrm{app},k}^{(\ell)})^\top\Delta_i^{\mathrm{phys}}\rVert_2^2}
{\lVert\Delta_i^{\mathrm{phys}}\rVert_2^2}.
\]
Five-fold scene cross-fitting excludes evaluation scenes when fitting $U_{\mathrm{app},k}$; dimension-matched Haar-random subspaces provide the reference. The primary $k=4$ was retained from development before sealed effects were observed. Every unrelated pair uses different scenes, possible-to-impossible orientation is fixed, and the reported intervals use 10,000 scene-bootstrap draws. Shuffled-correspondence and random-subspace references each use 10,000 draws. Principal-angle diagnostics follow standard subspace comparison \citep{bjorck1973angles}; tiny-displacement, raw-feature, local-normalization, and anisotropy checks remain supplementary.}

\begin{table}[htbp]
\centering
\color{black}
\small
\setlength{\tabcolsep}{2.4pt}
\caption{Complete sealed magnitude and cross-fitted projection curves. Magnitude cells are hue/brightness/recolor/clean-physical, each divided by $G_\ell$. Projection is $P_{\mathrm{app},4}$ with a dimension-matched random mean near 0.00284 at every layer; brackets are 95\% scene-bootstrap intervals.}
\label{tab:layerwise-endpoints}
\begin{tabular}{@{}llrcc@{}}
\toprule
Model & Layer & $G_\ell$ & Relative magnitude H/B/R/P & $P_{\mathrm{app},4}$ [95\% CI] \\
\midrule
V-JEPA2-G & b00 & .607285 & .571/2.111/.610/.007 & .401 [.320, .483] \\
 & b10 & .007055 & .392/1.634/.449/.007 & .125 [.078, .178] \\
 & b20 & .006974 & .260/1.022/.245/.010 & .063 [.039, .090] \\
 & b30 & .015104 & .110/1.497/.142/.012 & .102 [.065, .145] \\
 & final & .235825 & .080/.077/.090/.021 & .060 [.050, .071] \\
\midrule
VideoMAE2-G & b00 & .010641 & 1.245/1.750/1.310/.005 & .348 [.268, .432] \\
 & b10 & .017612 & .347/.542/.366/.005 & .057 [.036, .083] \\
 & b20 & .031948 & .195/.291/.226/.008 & .061 [.050, .076] \\
 & b30 & .028516 & .137/.258/.161/.007 & .060 [.048, .076] \\
 & final & .057018 & .034/.115/.065/.006 & .077 [.063, .092] \\
\bottomrule
\end{tabular}
\end{table}

\begin{table}[htbp]
\centering
\color{black}
\small
\setlength{\tabcolsep}{3.5pt}
\caption{Development-to-sealed trajectory classification. Metrics and labels were fixed before the completed sealed geometry run.}
\label{tab:geometry-replication}
\begin{tabular}{@{}lcc@{}}
\toprule
Property & V-JEPA2-G & VideoMAE2-G \\
\midrule
Relative appearance magnitude & strong replication & strong replication \\
Null-adjusted direction & strong replication & strong replication \\
Appearance-physics overlap & qualitative replication & strong replication \\
\bottomrule
\end{tabular}
\end{table}

\begin{table}[htbp]
\centering
\color{black}
\small
\setlength{\tabcolsep}{2.5pt}
\caption{Final-layer sealed projection sensitivity. Cells report observed $P_{\mathrm{app},k}$ [95\% scene-bootstrap interval] / random mean. The reported $k=4$ was not changed.}
\label{tab:geometry-k-sensitivity}
\begin{tabular}{@{}lcccc@{}}
\toprule
Model & $k=1$ & $k=2$ & $k=4$ & $k=8$ \\
\midrule
V-JEPA2-G & .005 [.004,.006]/.0007 & .012 [.010,.014]/.0014 & .060 [.050,.071]/.0028 & .153 [.132,.174]/.0057 \\
VideoMAE2-G & .024 [.016,.034]/.0007 & .058 [.046,.073]/.0014 & .077 [.063,.092]/.0028 & .130 [.113,.148]/.0057 \\
\bottomrule
\end{tabular}
\end{table}

{\color{black}\paragraph{Directional robustness.} Raw MRL is insufficient: the shuffled V-JEPA2-G block-00 brightness population has first-PC energy 0.886, showing strong background anisotropy. Figure~\ref{fig:layerwise-structure} therefore reports $\Delta$MRL; Table~\ref{tab:layerwise-direction} gives full curves. Major observed MRL effects are not driven by one scene: leave-one-scene-out ranges are [0.781, 0.797] for V-JEPA2-G hue at block 30 (full 0.784), [0.953, 0.961] for brightness at block 30 (0.954), and [0.708, 0.720] for recoloring at block 30 (0.712). Final projection is likewise stable at 0.060 with leave-one-scene-out range [0.057, 0.060].}

\begin{table}[htbp]
\centering
\color{black}
\small
\setlength{\tabcolsep}{2.5pt}
\caption{Complete sealed directional curves. Each cell is observed MRL / shuffled-correspondence mean / difference. H, B, R, and P denote hue, brightness, selective recoloring, and clean possible-to-impossible displacement. MRL intervals and empirical permutation values are in the supplementary tables.}
\label{tab:layerwise-direction}
\begin{tabular}{@{}llcccc@{}}
\toprule
Model & Layer & Hue & Brightness & Recolor & Physics \\
\midrule
V-JEPA2-G & b00 & .651/.466/.186 & .835/.785/.050 & .427/.420/.007 & .065/.066/$-$.002 \\
 & b10 & .714/.384/.329 & .953/.741/.212 & .499/.330/.168 & .064/.066/$-$.002 \\
 & b20 & .758/.349/.409 & .941/.660/.281 & .615/.282/.333 & .088/.066/.022 \\
 & b30 & .784/.250/.534 & .954/.699/.255 & .712/.250/.462 & .119/.066/.052 \\
 & final & .693/.212/.481 & .792/.236/.556 & .564/.180/.383 & .134/.067/.067 \\
\midrule
VideoMAE2-G & b00 & .666/.499/.167 & .962/.763/.199 & .449/.461/$-$.012 & .070/.067/.003 \\
 & b10 & .720/.380/.340 & .947/.578/.369 & .578/.323/.254 & .079/.065/.014 \\
 & b20 & .742/.310/.432 & .942/.465/.477 & .658/.283/.375 & .069/.065/.004 \\
 & b30 & .747/.270/.477 & .955/.450/.506 & .693/.257/.436 & .078/.065/.012 \\
 & final & .704/.139/.566 & .965/.327/.637 & .783/.191/.592 & .075/.065/.010 \\
\bottomrule
\end{tabular}
\end{table}

The final-layer transformed-to-clean nearest-feature retrieval rates are 65.6\% for V-JEPA2-G and 44.6\% for VideoMAE2-G. These measure recovery of source-video identity under the evaluated transformations. Appearance accessibility is evaluated separately by the linear probes.

\subsection{{\color{black}Exploratory transformation trajectories (run after the sealed test)}}
\label{app:multidose-geometry}

{\color{black}\textbf{Status: exploratory, run after the sealed test.} This analysis does not change the sealed geometry result. It applies the exact paper V-JEPA2-G checkpoint to corrected development (38 scenes, 76 pairs, 152 source videos) at the five reported depths. Each source has clean video plus hue rotations of $15^\circ$, $45^\circ$, and $90^\circ$; brightness multipliers 0.85, 0.65, and 0.45; and selective recoloring rotations of $45^\circ$, $90^\circ$, and $180^\circ$. Brightness strength is $1-b$. Scene is the independent unit and intervals use 10,000 bootstrap draws. We use these trajectories to probe transformation geometry and local equivariance descriptively \citep{lenc2015equivariance}.}

{\color{black}For an ordered trajectory $z(a_0),\ldots,z(a_m)$, straightness is endpoint distance divided by path length, $\lVert z(a_m)-z(a_0)\rVert_2/\sum_j\lVert z(a_{j+1})-z(a_j)\rVert_2$. Displacement dimensionality uses entropy effective rank, $\exp[-\sum_j p_j\log p_j]$, where $p_j$ is normalized squared singular-value energy. These quantities describe ray-likeness and dimensionality; neither identifies a semantic axis.}

\begin{figure}[htbp]
\centering
\includegraphics[width=\linewidth]{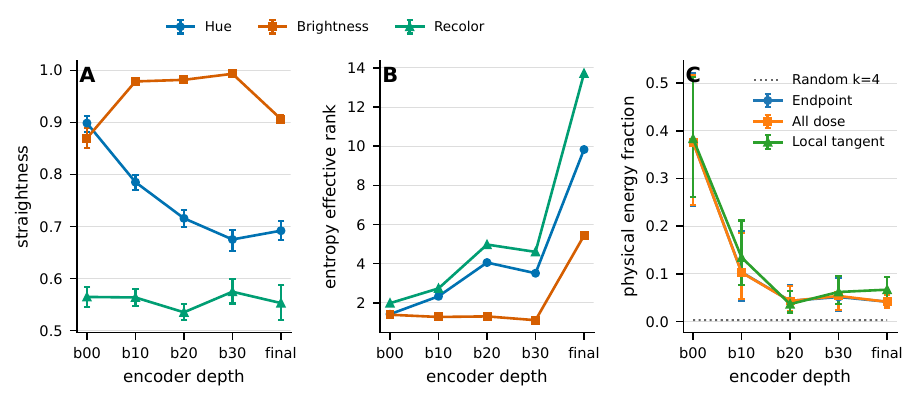}
\caption{{\color{black}V-JEPA2-G geometry across transformation doses on the 38 development scenes. (A) Scene-mean trajectory straightness with 95\% scene-bootstrap intervals. (B) Entropy effective rank of the pooled scene-dose displacement matrix. (C) Five-fold held-out physical projection into strongest-endpoint, all-dose, and local-tangent appearance spaces, all matched at $k=4$, with scene-bootstrap intervals and a dimension-matched random reference. The tangent advantage is layer dependent and does not replace the frozen sealed projection result.}}
\label{fig:exploratory-trajectories}
\end{figure}

{\color{black}Figure~\ref{fig:exploratory-trajectories} shows that at final, brightness is the most ray-like evaluated factor: straightness is 0.907 [0.899, 0.914], versus 0.692 [0.673, 0.710] for hue and 0.553 [0.520, 0.587] for selective recoloring. Final effective ranks are 5.43, 9.83, and 13.72, respectively. Thus selective recoloring is higher-dimensional than the other evaluated factors, while brightness remains only \emph{relatively} ray-like and has more than one effective dimension.}

{\color{black}For hue, held-out relative error is 0.798 for a shared-ray model and 0.747 for a harmonic curved model. This is modest evidence that the observed $0^\circ$--$90^\circ$ arc is better represented by a curved model than one common ray, but the final effective rank of 9.83 makes the low-dimensional part of the hypothesis only partial. Three nonzero doses spanning at most $90^\circ$ cannot test closure, periodicity, circularity, or a latent color wheel.}

{\color{black}The main exploratory result compares a local appearance basis with an endpoint basis of the same dimension. At final and $k=4$, held-out physical projection is 0.0416 [0.0297, 0.0559] into the strongest-endpoint appearance space, 0.0414 [0.0295, 0.0564] into the all-dose space, and 0.0668 [0.0448, 0.0932] into the local appearance-tangent space; the random reference is approximately 0.00284. The paired tangent-minus-endpoint difference is 0.0252 [0.0072, 0.0450]. The advantage is not universal across depth: at block 20, tangent projection (0.037) is below endpoint projection (0.043). The final result therefore says only that a matched-$k$ local appearance-tangent basis captures more held-out physical-displacement energy than the strongest-endpoint basis at that layer; it does not establish entanglement, causal factorization, or physical understanding.}

\begin{table}[htbp]
\centering
\color{black}
\small
\caption{Qualitative V-JEPA2-L pilot to exact V-JEPA2-G replication check at the final representation. The models differ in scale and the G run uses the corrected 38-scene development population.}
\label{tab:multidose-lg}
\begin{tabular}{@{}lrrrr@{}}
\toprule
Factor & L straightness & G straightness & L effective rank & G effective rank \\
\midrule
Hue & .739 & .692 & 3.61 & 9.83 \\
Brightness & .887 & .907 & 2.98 & 5.43 \\
Selective recolor & .324 & .553 & 5.94 & 13.72 \\
\bottomrule
\end{tabular}
\end{table}

{\color{black}Table~\ref{tab:multidose-lg} gives the qualitative classifications: factor-specific geometry replicated; brightness as the most ray-like factor replicated; recoloring as higher-dimensional replicated; hue as curved and low-dimensional replicated only partially; and final tangent projection exceeding endpoint projection replicated. Table~\ref{tab:multidose-direction} reports direction by dose. These labels are a cross-check and say nothing about the effect of model scale.}

\begin{table}[htbp]
\centering
\color{black}
\scriptsize
\setlength{\tabcolsep}{4pt}
\caption{Final-layer direction by dose. Each row is observed MRL / shuffled mean / $\Delta$MRL.}
\label{tab:multidose-direction}
\begin{tabular}{@{}lccc@{}}
\toprule
Factor & Mild dose & Medium dose & Strong dose \\
\midrule
Hue ($15^\circ/45^\circ/90^\circ$) & .640/.064/.576 & .685/.160/.525 & .734/.234/.500 \\
Brightness (.85/.65/.45) & .810/.081/.729 & .830/.164/.667 & .816/.257/.559 \\
Recolor ($45^\circ/90^\circ/180^\circ$) & .598/.138/.459 & .636/.195/.442 & .625/.213/.412 \\
\bottomrule
\end{tabular}
\end{table}
\FloatBarrier

\section{Action and appearance controls}
\subsection{Action mismatch and hue controls}
\label{app:additional-protocols}

\subsubsection{{\color{black}LeWorldModel, DINO-WM, and JEPA-WM controls}}
\label{app:lewm-dinowm-controls}
The TwoRoom and PushT studies strict-load the released LeWorldModel checkpoints, preserve the official four-frame, frame-skip-five preprocessing and action normalization, and use 256 fresh independent episodes per task. LeWorldModel combines latent prediction with the SIGReg regularizer introduced in LeJEPA \citep{balestriero2025lejepa}. For each episode we simulate lawful targets from the third context state under scaled five-action blocks $ca$ for $c\in\{0,0.25,0.5,0.75,1\}$, render the lawful target once, and reuse the exact same target bytes under lawful $ca$, zero, and opposite $-ca$ action blocks. The LeWM native score is mean squared error between predicted and encoded target latents. The DINO-WM PushT extension uses the same 256 PushT episodes, target bytes, action blocks, and rendered-effect matches; it normalizes actions and proprioception with the released DINO-WM PushT statistics and scores final-step visual-latent MSE between the predictor output and encoded target. {\color{black}The JEPA-WM PushT extension uses the same controlled construction and three-frame teacher-forcing context with the released 224-pixel preprocessing, action and proprioception statistics, AdaLN plus RoPE predictor, and final one-step visual embedding MSE over DINOv2 patch tokens. The visual score excludes proprioceptive prediction loss, and the analysis is post-hoc exploratory.} Action contrasts are same-target differences relative to the lawful-action baseline. The reported hue-minus-opposite contrasts instead subtract the opposite-action score for a moving target from the hue-modified static-target score. They are absolute score comparisons across different target baselines, with predeclared rendered-effect matches. Episode-level 95\% intervals use 10,000 bootstrap draws over episodes. Appendix Sections~\ref{app:pusht-role-design}--\ref{app:pusht-role-decomposition} separately test matched color directions and decompose their latent source.

\begin{table}[htbp]
\centering
\scriptsize
\setlength{\tabcolsep}{2.4pt}
\caption{Complete native prediction-error contrasts (256 episodes per task or scorer; 95\% episode-bootstrap intervals). LeWM rows use native prediction MSE; DINO-WM PushT rows use final-step visual-latent MSE. {\color{black}JEPA-WM PushT rows use final one-step visual embedding MSE and are post-hoc exploratory. Native magnitudes are not comparable across models.} Action rows subtract the lawful score for the same target; hue rows are absolute score differences between the hue-modified static target and the matched opposite-action condition. They do not compare the changes the two interventions induce.}
\label{tab:native-error-full}
\begin{tabular}{@{}>{\raggedright\arraybackslash}p{0.16\linewidth}>{\raggedright\arraybackslash}p{0.20\linewidth}>{\raggedright\arraybackslash}p{0.302\linewidth}>{\raggedright\arraybackslash}p{0.302\linewidth}@{}}
\toprule
Model/task & Contrast & Low or matched scale & High or matched scale \\
\midrule
{\leavevmode\color{black}LeWM TwoRoom} & {\leavevmode\color{black}zero vs. lawful} & 0.359 [0.315, 0.403] at 0.25 & 1.463 [1.382, 1.545] at 1.0 \\
{\leavevmode\color{black}LeWM TwoRoom} & {\leavevmode\color{black}opposite vs. lawful} & 0.909 [0.834, 0.987] at 0.25 & 1.884 [1.821, 1.946] at 1.0 \\
{\leavevmode\color{black}LeWM TwoRoom} & {\leavevmode\color{black}hue vs. opposite} & 0.222 [0.150, 0.296] at 120$^\circ$/0.75 & 0.220 [0.147, 0.295] at 180$^\circ$/1.0 \\
\midrule
LeWM PushT & {\leavevmode\color{black}zero vs. lawful} & 0.0085 [0.0070, 0.0102] at 0.25 & 0.1668 [0.1469, 0.1880] at 1.0 \\
LeWM PushT & {\leavevmode\color{black}opposite vs. lawful} & 0.0332 [0.0279, 0.0391] at 0.25 & 0.3834 [0.3456, 0.4242] at 1.0 \\
LeWM PushT & {\leavevmode\color{black}hue vs. opposite} & 0.146 [0.125, 0.167] at 30$^\circ$/0.25 & 0.642 [0.585, 0.696] at 60$^\circ$/0.75 \\
\midrule
DINO-WM PushT & {\leavevmode\color{black}zero vs. lawful} & 0.00052 [0.00046, 0.00059] at 0.25 & 0.00183 [0.00152, 0.00215] at 1.0 \\
DINO-WM PushT & {\leavevmode\color{black}opposite vs. lawful} & 0.00104 [0.00091, 0.00118] at 0.25 & 0.00364 [0.00304, 0.00428] at 1.0 \\
DINO-WM PushT & {\leavevmode\color{black}hue vs. opposite} & $-$0.03082 [$-$0.04028, $-$0.02273] at 30$^\circ$/0.25 & $-$0.10987 [$-$0.12416, $-$0.09614] at 60$^\circ$/0.75 \\
\midrule
{\leavevmode\color{black}JEPA-WM PushT} & {\leavevmode\color{black}zero vs. lawful} & {\leavevmode\color{black}0.00049 [0.00042, 0.00056] at 0.25} & {\leavevmode\color{black}0.00179 [0.00151, 0.00209] at 1.0} \\
{\leavevmode\color{black}JEPA-WM PushT} & {\leavevmode\color{black}opposite vs. lawful} & {\leavevmode\color{black}0.00098 [0.00084, 0.00112] at 0.25} & {\leavevmode\color{black}0.00351 [0.00293, 0.00411] at 1.0} \\
{\leavevmode\color{black}JEPA-WM PushT} & {\leavevmode\color{black}hue vs. opposite} & {\leavevmode\color{black}$-$0.03670 [$-$0.04434, $-$0.03008] at 30$^\circ$/0.25} & {\leavevmode\color{black}$-$0.12481 [$-$0.13905, $-$0.11114] at 60$^\circ$/0.75} \\
\bottomrule
\end{tabular}
\end{table}

\subsubsection{Separating target difficulty from intervention effects}
\label{app:pusht-baselines}
Let $S_0$ denote the static lawful score, $S_h$ the hue-modified static score, $S_l$ the moving lawful score, and $S_o$ the score for that same moving target under the opposite action. The reported cross-target contrast decomposes as
\[
S_h-S_o=(S_h-S_0)-(S_o-S_l)-(S_l-S_0).
\]
The first two terms are the hue and action increments relative to their respective lawful baselines. The third is the difference in lawful target difficulty. Table~\ref{tab:pusht-baseline-means} gives the stored condition means at the higher PushT rendered-effect match. It clarifies the existing score comparison without introducing a new inferential test.

\begin{table}[htbp]
\centering
\small
\caption{Stored PushT condition means at hue $60^\circ$ and action scale 0.75 (256 episodes per model). Increments are differences of these means; no new paired intervals are estimated here. All values are native latent MSE and are interpreted within each model.}
\label{tab:pusht-baseline-means}
\begin{tabular}{@{}lrrrrrr@{}}
\toprule
Model & $S_0$ & $S_h$ & $S_l$ & $S_o$ & $S_h-S_0$ & $S_o-S_l$ \\
\midrule
LeWM & 0.008728 & 0.895323 & 0.008358 & 0.253505 & 0.886596 & 0.245146 \\
DINO-WM & 0.061985 & 0.111752 & 0.218553 & 0.221626 & 0.049767 & 0.003073 \\
JEPA-WM & 0.196626 & 0.214008 & 0.335983 & 0.338820 & 0.017381 & 0.002837 \\
\bottomrule
\end{tabular}
\end{table}

\begin{figure}[htbp]
\centering
\includegraphics[width=\linewidth]{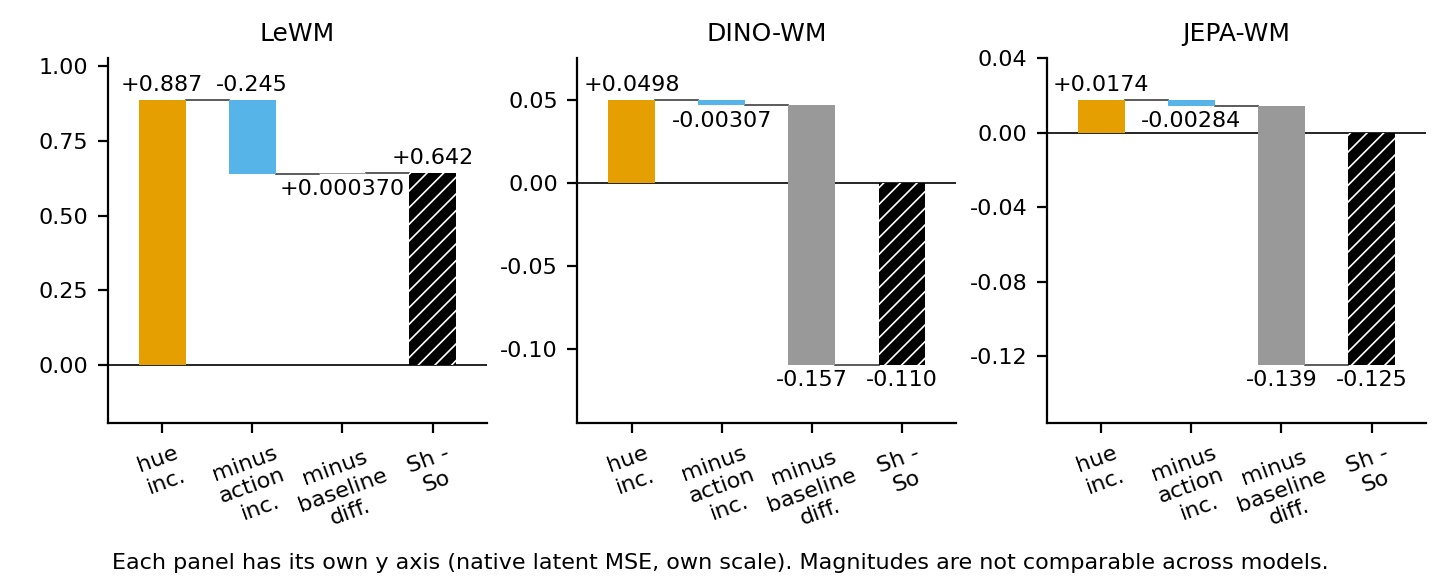}
\caption{Descriptive decomposition of the stored PushT means in Table~\ref{tab:pusht-baseline-means}, at hue $60^\circ$ and action scale 0.75. The action increment and moving-minus-static lawful baseline difference are subtracted from the hue increment to obtain $S_h-S_o$. Each model uses its own native latent-MSE scale.}
\label{fig:pusht-baseline-waterfall}
\end{figure}

The mean hue increment exceeds the mean action increment for all three models at this match. For the DINO-based models, however, the moving lawful baseline is much higher than the static baseline, producing a negative absolute $S_h-S_o$ contrast. The sign of that contrast therefore cannot identify which intervention changes the score more. Neither contrast makes the two interventions equally predictable, equally relevant to the task, or equally far from the training distribution. The shared DINOv2 encoder and differing training pipelines also prevent attributing the result to color augmentation alone.

\FloatBarrier
\subsection{PushT role-color decomposition}
\label{app:pusht-role-color}

\subsubsection{Question and matched design}
\label{app:pusht-role-design}
{\color{black}PushT uses blue for the agent, gray for the block, and green for the goal. Color is a potentially task-relevant cue in PushT.} We test whether native LeWorldModel prediction MSE depends only on the amount of rendered change or also on color direction. The frozen confirmatory run uses 256 independent episodes excluded from all earlier PushT pilots, the released checkpoint, official four-frame and frame-skip-five preprocessing, and the installed Pymunk/Pygame renderer. Because the installed renderer is close but not byte-identical to stored observations, every contrast uses the paired current-renderer output as its baseline.

\begin{figure}[htbp]
\centering
\includegraphics[width=\linewidth]{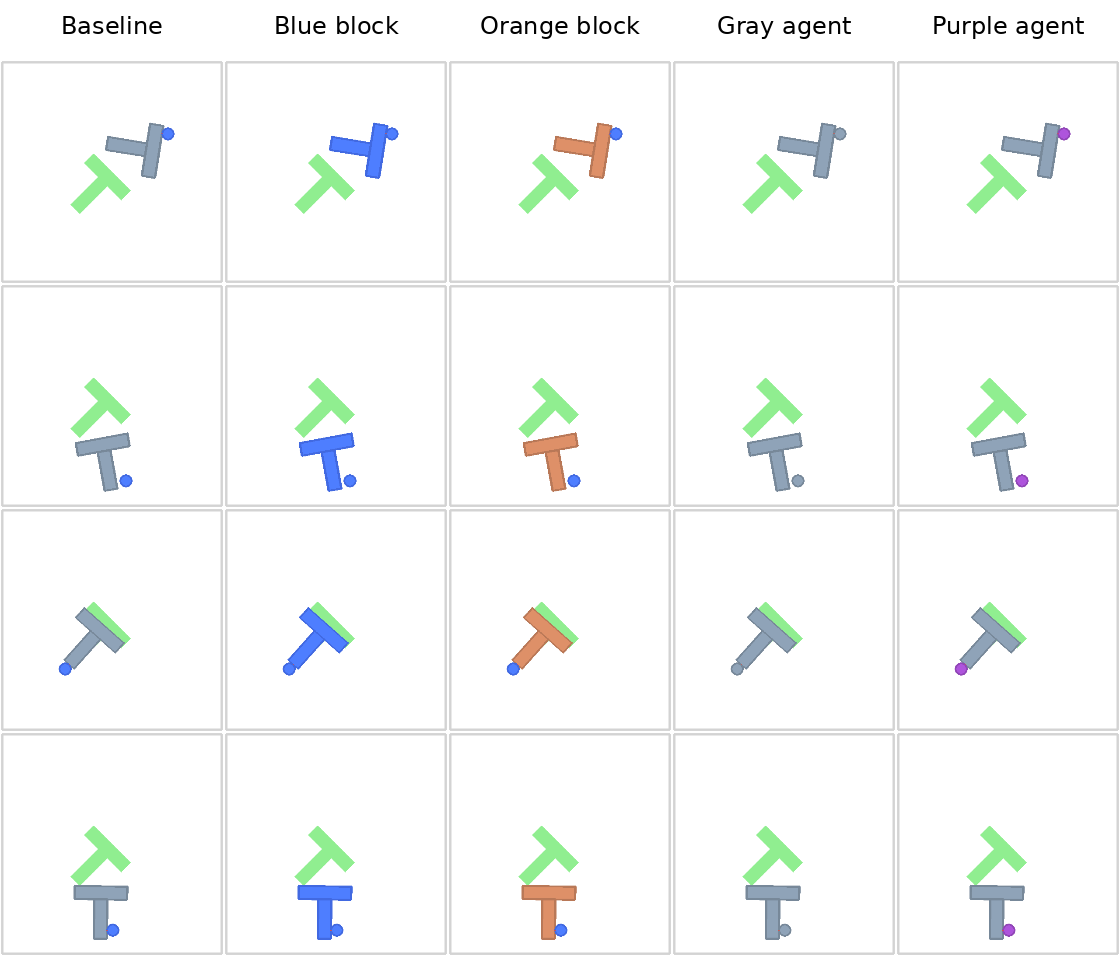}
\caption{Existing PushT role-color examples from the LeWorldModel study. Each row holds the simulator state fixed. Columns show the paired renderer baseline, an agent-blue block, a matched novel-orange block, a block-gray agent, and a matched novel-purple agent. These target-frame recolorings illustrate the conditions in Table~\ref{tab:role-color-design}; they do not change object poses or motion.}
\label{fig:pusht-role-examples}
\end{figure}

The block is recolored to agent blue and compared with a novel orange selected to match rendered magnitude; the agent is recolored to block gray and compared with a matched novel purple. Figure~\ref{fig:pusht-role-examples} illustrates these paired recolorings. Each condition is applied either only to the target or persistently to all four frames. Across colors, the maximum realized simulator-state difference is zero. In target-only conditions, context pixels, actions, and predictor outputs are exactly identical; only the target-encoder representation changes. Table~\ref{tab:role-color-design} summarizes the matched rendered changes.

\begin{table}[htbp]
\centering
\small
\setlength{\tabcolsep}{5pt}
\caption{Matched PushT recolorings. Values are episode means relative to the paired current-renderer baseline.}
\label{tab:role-color-design}
\begin{tabular}{llllcc}
\toprule
Object & Role color & Matched novel & Role/novel RGB L1 & Role/novel changed pixels \\
\midrule
Block & agent blue & orange & 0.006157/0.006188 & 0.02924/0.02924 \\
Agent & block gray & purple & 0.000563/0.000563 & 0.002844/0.002841 \\
\bottomrule
\end{tabular}
\end{table}

\subsubsection{Matched contrasts}
\label{app:pusht-role-contrasts}
For each object and scope, the contrast, declared before the run, is the score with the role color minus the score with the matched novel color. Episode-level 95\% intervals use 10,000 paired-bootstrap draws. All four directional contrasts are positive in the mean (Table~\ref{tab:role-color-contrasts}), including substantially matched target RGB change and affected-pixel fraction.

\begin{table}[htbp]
\centering
\small
\setlength{\tabcolsep}{2.4pt}
\caption{Role-color minus matched-novel prediction-MSE contrasts over 256 fresh episodes.}
\label{tab:role-color-contrasts}
\begin{tabular}{@{}>{\raggedright\arraybackslash}p{0.14\linewidth}>{\raggedright\arraybackslash}p{0.21\linewidth}>{\raggedright\arraybackslash}p{0.305\linewidth}>{\raggedright\arraybackslash}p{0.305\linewidth}@{}}
\toprule
Contrast & Scope & Mean [95\% CI] & Positive episodes \\
\midrule
Block blue $-$ orange & target only & 0.493 [0.450, 0.535] & 233/256 \\
Block blue $-$ orange & persistent & 0.111 [0.092, 0.131] & 191/256 \\
Agent gray $-$ purple & target only & 0.397 [0.355, 0.440] & 245/256 \\
Agent gray $-$ purple & persistent & 0.088 [0.054, 0.123] & 164/256 \\
\bottomrule
\end{tabular}
\end{table}

\subsubsection{Exact latent decomposition}
\label{app:pusht-role-decomposition}
Let $r_0=p_0-z_0$ be the baseline residual and $q=\Delta p-\Delta z$ the intervention-induced residual shift. Because the native score is mean squared latent error,
\[
\Delta S
=
2\,\operatorname{mean}(r_0\odot q)
+
\operatorname{mean}(q^2).
\]
Reconstructing every intervention score from this identity has maximum absolute error $2.38\times10^{-7}$. In target-only conditions $\Delta p=0$ exactly, and the role-minus-novel score gaps nearly equal the corresponding target-encoder shift gaps (Table~\ref{tab:role-color-decomposition}). This locates the abrupt effect primarily in target-encoder color-direction sensitivity.

\begin{table}[htbp]
\centering
\small
\setlength{\tabcolsep}{4pt}
\caption{Matched-color decomposition. Cosines compare persistent predictor and target shifts for the role-color and matched-novel conditions.}
\label{tab:role-color-decomposition}
\begin{tabular}{lcccc}
\toprule
Pair & Target score gap & Target-shift gap & Persistent score gap & Shift cosines \\
\midrule
Block blue/orange & 0.493 & 0.495 & 0.111 & 0.916/0.928 \\
Agent gray/purple & 0.397 & 0.400 & 0.088 & 0.815/0.705 \\
\bottomrule
\end{tabular}
\end{table}

Persistent predictor and target shifts point in similar directions, especially for block recoloring, but not identically (cosines 0.71 to 0.93 across the four conditions). The experiment therefore supports color-direction sensitivity and partial agreement between predictor and target shifts under persistent changes. 

\FloatBarrier

\section{Physion++ static and appearance diagnostics}
\label{app:physionpp}
\subsection{Full-population controls}
\label{app:physion-full}
The fixed official Mass manifests contain 192 videos / 96 pairs: 32 pairs each in Mass-Collision, Mass-Dominoes, and Mass-Waterpush. Additional directory files do not enlarge this cohort. Every complete nonoverlapping 16-frame window is used for the static study (5,177 windows); incomplete tails are recorded and left unpadded. Each window is scored as original and with frame 0, 8, or 15 repeated throughout. The appearance study selects window $\lfloor n/2\rfloor$ from each video's $n$ complete windows, without using scores, and evaluates original, persistent hue, target-only hue, persistent darkening, and target-only darkening. Hue adds 21 PIL HSV bins modulo 256; darkening applies $\lfloor4\,\mathrm{RGB}/5\rfloor$. Persistent transformations affect all 16 frames and target-only transformations frames 12-15. Paired future pixels/tensors match exactly; target-only context pixels/tensors match original.

The exact archived scoring body uses FP16 CUDA batch-one evaluation, 12 selected context frames / four target frames, FP32 layer normalization of both operands, and mean L1 over 512 target tokens.

\textbf{The Transformers implementation encodes all 16 frames before selecting context tokens.} Thus even target-only image changes can alter the predictor; identical future pixels under persistent/target-only transformations need not yield identical target embeddings. This diagnostic concerns the released full-clip scoring implementation, not a causal forecast conditioned only on the first 12 frames. We kept this implementation after seeing the results and did not replace it.

For static results, let $d_{vw}$ be original minus the mean of the three static scores in window $w$ of video $v$. We first average $d_{vw}$, or $\mathbb{1}[d_{vw}\leq0]$, within each video, then average the two videos within each matched pair. This is not an independent-window prevalence estimate. We use 50,000 scenario-stratified pair-bootstrap draws, seed 2026102601; layout-cluster sensitivity uses seed 2026102701 and 13/10/19 layout clusters. The mean residual is $-0.028264$ with pair 95\% interval [$-0.028999$,$-0.027525$] and layout 95\% interval [$-0.029265$,$-0.027421$]. Scenario means are $-0.032012$, $-0.031993$, and $-0.020788$, respectively. Mean within-video nonpositive fractions are 89.20\%, 88.01\%, and 76.17\%; their overall mean is 84.46\% [pair 95\% interval 83.67\%, 85.24\%], not a physical-violation false-positive rate.

Appearance contrasts also average the two videos within each pair. We use 50,000 scenario-stratified draws, seed 2026102602, and 99.1667\% intervals for the six fixed contrasts (familywise Bonferroni level 0.05); layout sensitivity uses seed 2026102702. Main Table~\ref{tab:physion-appearance} reports all six. Layout-adjusted intervals retain negative persistent effects and negative persistent-minus-target-only effects; target-only hue is sensitive to the resampling unit (layout interval [$-0.001676$,$-0.000012$], while the primary pair interval crosses zero). Table~\ref{tab:physion-scenarios} gives every scenario mean, including sign reversals; none was used to select the population or transformation.
\begin{table}[htbp]
\centering
\small
\setlength{\tabcolsep}{3pt}
\caption{All scenario-specific Physion++ appearance contrasts (means; 32 matched pairs per scenario). Overall intervals are reported in the main text; scenario rows are descriptive, not separately selected tests.}
\label{tab:physion-scenarios}
\begin{tabular}{@{}lrrr@{}}
\toprule
Contrast & Collision & Dominoes & Waterpush \\
\midrule
Persistent hue $-$ original & $-0.001965$ & $-0.003126$ & $-0.004274$ \\
Target-only hue $-$ original & $-0.002452$ & $+0.001806$ & $-0.001433$ \\
Persistent darkening $-$ original & $-0.004205$ & $-0.000940$ & $-0.002966$ \\
Target-only darkening $-$ original & $-0.002640$ & $+0.005379$ & $-0.000978$ \\
Persistent $-$ target-only hue & $+0.000487$ & $-0.004932$ & $-0.002841$ \\
Persistent $-$ target-only darkening & $-0.001565$ & $-0.006318$ & $-0.001989$ \\
\bottomrule
\end{tabular}
\end{table}
\begin{table}[htbp]
\centering
\small
\setlength{\tabcolsep}{3pt}
\caption{Descriptive FP64 hybrid-L1 contributions averaged over all 192 videos (equally paired). Target and predictor contributions sum to the score change, up to rounding; they are not causal effects.}
\label{tab:physion-hybrids}
\begin{tabular}{@{}lrrr@{}}
\toprule
Condition & Score change & Target contribution & Predictor contribution \\
\midrule
persistent hue & $-0.003122$ & $-0.000825$ & $-0.002296$ \\
target only hue & $-0.000693$ & $+0.002942$ & $-0.003635$ \\
persistent brightness & $-0.002704$ & $+0.001699$ & $-0.004403$ \\
target only brightness & $+0.000587$ & $+0.002281$ & $-0.001694$ \\
\bottomrule
\end{tabular}
\end{table}

For normalized predictor $p$ and target $t$, define $E_{ij}=\operatorname{mean}|p_i-t_j|$ for original ($0$) and transformed ($1$) clips. Descriptive contributions are
\[
C_t=\tfrac12[(E_{01}-E_{00})+(E_{11}-E_{10})],\qquad
C_p=\tfrac12[(E_{10}-E_{00})+(E_{11}-E_{01})].
\]
They sum algebraically to $E_{11}-E_{00}$. FP64 hybrid checks use $10^{-10}$ identity tolerance and $10^{-6}$ native reconstruction tolerance; no intervention fixes one operand in the actual network. Table~\ref{tab:physion-hybrids} is therefore bookkeeping, not causal mediation. In particular, target-only predictor contributions need not vanish. Pixel darkening is not physical relighting, and contact labels do not label violations.

\FloatBarrier
\subsection{Historical static-audit selection and scoring}
\label{app:physion-static-details}
\label{app:physion-selection}
The checkpoint identity is recorded in the \href{https://github.com/3-ml/jepa-surprise/blob/camera-ready-2026/docs/reproducibility.md#checkpoint-and-model-identities}{online checkpoint table}.
Ten additional Physion++ Mass test videos were selected deterministically as evenly spaced files after excluding the earlier single-video debug example. Every complete 16-frame segment was scored, yielding 273 segments. For each segment we record the native latent residual, mean consecutive absolute RGB difference on the 0-255 scale, and the fraction of pixels whose consecutive-frame change exceeds 5.

For static calibration, the first, middle, or last decoded frame is repeated across all 16 input positions and rescored. The first-frame event residual is the original score minus the corresponding frozen-frame score. This control does not estimate a universal corrected surprise score; it asks only whether a local raw peak requires temporal change.

\subsection{Motion association and static candidates}
\label{app:physion-results}
Across all segments, raw score and mean consecutive RGB change have Pearson correlation $r=0.596$. Each video's highest raw-score segment contains substantial motion; their mean pixel-motion value is 3.712 and their first-frame residuals are positive, from 0.0165 to 0.0492. However, 6 of 10 videos also contain a high-scoring near-static segment with mean RGB change below 0.01. Table~\ref{tab:physion-static-candidates} shows that all six have nonpositive frozen-frame residuals.

\begin{table}[htbp]
\centering
\small
\setlength{\tabcolsep}{5pt}
\caption{Highest-scoring near-static candidate in each affected Physion++ video. RGB change is on the 0-255 scale; residual is original minus first-frame-frozen score.}
\label{tab:physion-static-candidates}
\begin{tabular}{rrrrr}
\toprule
Video & Segment & Raw score & Mean RGB change & Static residual \\
\midrule
0 & 12 & 0.5898 & 0.001867 & $-0.0488$ \\
1 & 9  & 0.6021 & 0.001294 & $-0.0400$ \\
2 & 11 & 0.6083 & 0.009740 & $-0.0313$ \\
3 & 21 & 0.6380 & 0.005892 & $-0.0267$ \\
4 & 16 & 0.6215 & 0.001611 & $-0.0367$ \\
5 & 20 & 0.6455 & 0.001180 & $-0.0266$ \\
\bottomrule
\end{tabular}
\end{table}

The mean first-frame residual over these six candidates is $-0.0350$. The threshold and highest-candidate rule were diagnostic choices, so 6/10 is not a population false-positive rate. Raw V-JEPA2-L error partially tracks visible motion but also admits high-error static-content candidates. This motivates the main paper's pixel-motion and static baselines; it is not evidence about V-JEPA2-G benchmark accuracy or a general event detector.
\FloatBarrier

\section{Physical-validity tests with Kubric}
\label{app:final256}

\subsection{Population, rendering, and native score}
\begin{figure}[htbp]
\centering
\includegraphics[width=\linewidth]{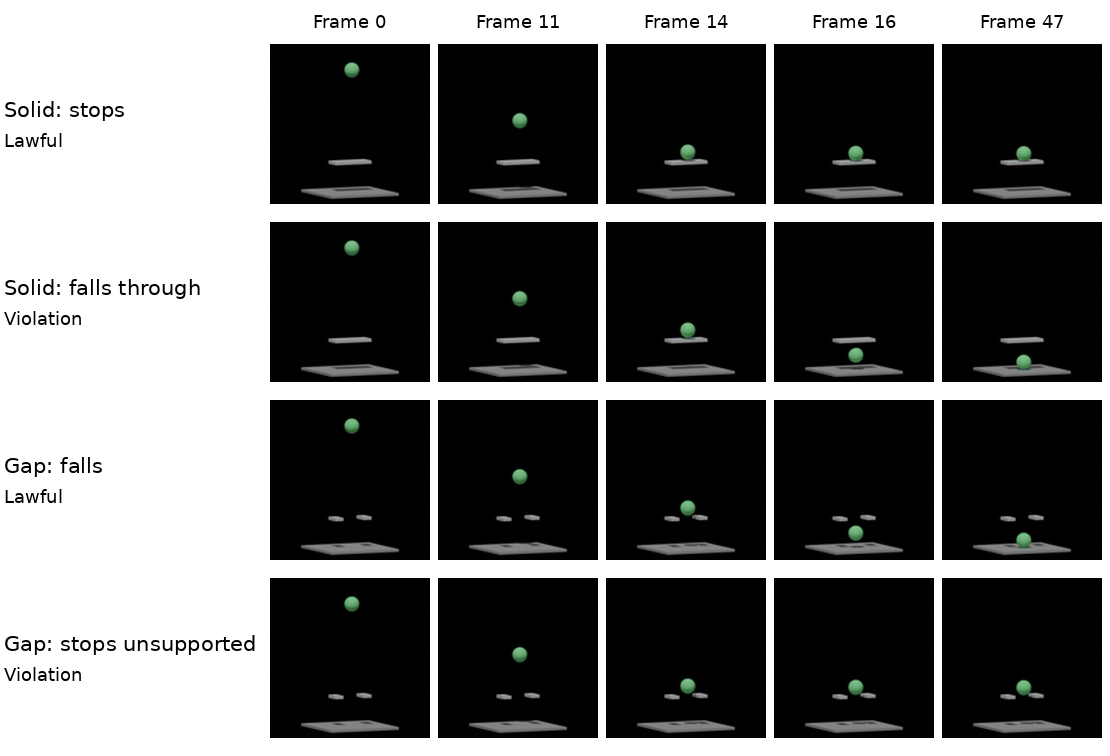}
\caption{Rendered illustration of the crossed-support construction. The small platform at middle height is either solid or split around a gap; the lower platform catches the falling sphere. The same stopping trajectory is lawful on solid support and unsupported over the gap. The same falling trajectory passes unlawfully through the solid platform but lawfully through the gap. Frames are cropped from the archived clean contact sheets for the first of the 24 development configurations (\texttt{dev24-001} in the released records), not selected from the final test. Its native impossible-minus-possible margins are $-0.00345$ (solid) and $+0.00830$ (gap); Figure~\ref{fig:final256} reports the independent 256-configuration evaluation.}
\label{fig:support-examples}
\end{figure}
After development on 24 configurations (288 clips), we froze 256 candidates with seeds 2026102700--2026102955 before rendering or scoring. Each candidate independently samples sphere radius in $[0.60,0.70]$, nominal contact time in $[1.18,1.24]$, initial horizontal position in $[-0.035,0.035]$, one of four camera positions, and one of four development colors. Initial height follows the specified free-fall contact time. The manifest fixes all draws. Each configuration crosses solid/gap support with lawful/inconsistent stopping or falling and clean/persistent/abrupt $30^\circ$ hue, giving 12 clips. Appearance does not change the simulated trajectory. Abrupt hue begins at frame 12; persistent hue applies throughout. Support is simulator-fixed. Across worlds the stop trajectories match, as do the fall trajectories, but support appearance, shadows, and occlusion differ.

Clips contain 48 frames at 12 fps, with 240-Hz simulation and $256\times256$ rendering (Cycles CUDA, 64 samples/pixel, no denoising). The scorer uses 12 context frames and 36 future frames (18 target tubelets), FP32 weights, bfloat16 compute/subtraction, and FP32 mean absolute error against normalized EMA-target tokens.

This context-only predictor path is distinct from the full-clip V-JEPA2-L path in Appendix~\ref{app:physion-full}.

\subsection{Estimands, threshold, and audit disposition}
\label{app:final256-estimands}
The primary family comprises four means of inconsistent minus lawful error: solid/gap under clean/persistent hue. We resample all conditions within complete configurations together, with 50,000 draws (seed 202610256) and 98.75\% percentile intervals, Bonferroni-adjusted for four means. Bootstrap coverage is approximate. A consistently positive ordering would require all four intervals to lie above zero. The two solid-support intervals lie below zero instead. The $0.001$ planning target denotes reporting resolution, not an equivalence or useful-effect margin.

The four inconsistent-minus-lawful means and their 98.75\% intervals are: clean solid $-0.006734$ [$-0.007421$, $-0.006045$]; clean gap $+0.006908$ [$+0.006189$, $+0.007632$]; persistent solid $-0.007839$ [$-0.008446$, $-0.007233$]; and persistent gap $+0.007495$ [$+0.006897$, $+0.008095$]. These are the complete numerical counterparts of Figure~\ref{fig:final256}'s ordering panel.

The threshold $0.4858863055706024$ was fitted on the 24 clean development configurations by maximum balanced accuracy with largest-threshold tie breaking; flagging uses $\geq$. No final refit, score inversion, or world-specific threshold is allowed. Secondary rates and AUROC use 10,000 configuration-bootstrap draws and descriptive 95\% intervals. Table~\ref{tab:final256-thresholds} reports both worlds separately. The balanced rows average the two worlds; they are not an independent sample or an equivalence result.
\begin{table}[htbp]
\centering
\small
\setlength{\tabcolsep}{3pt}
\caption{Unchanged development threshold on final256. Rates and accuracy are percentages; accuracy brackets are descriptive 95\% configuration-bootstrap intervals. Balanced rows weight the two worlds equally. AUROC is descriptive; the native score direction is not inverted.}
\label{tab:final256-thresholds}
\begin{tabular}{@{}lrrrr@{}}
\toprule
Appearance / world & False alarms & Misses & Balanced accuracy & AUROC \\
\midrule
Clean / Solid & 100.00 & 45.70 & 27.15 [24.02, 30.27] & 0.083 \\
Clean / Gap & 37.89 & 0.78 & 80.66 [77.54, 83.59] & 0.882 \\
Clean / Balanced & 68.95 & 23.24 & 53.91 [52.05, 55.76] & 0.499 \\
Persistent / Solid & 100.00 & 59.38 & 20.31 [17.38, 23.44] & 0.048 \\
Persistent / Gap & 30.47 & 0.39 & 84.57 [81.64, 87.50] & 0.944 \\
Persistent / Balanced & 65.23 & 29.88 & 52.44 [50.68, 54.20] & 0.492 \\
Abrupt / Solid & 100.00 & 65.62 & 17.19 [14.26, 20.12] & 0.042 \\
Abrupt / Gap & 28.52 & 0.00 & 85.74 [82.81, 88.48] & 0.948 \\
Abrupt / Balanced & 64.26 & 32.81 & 51.46 [49.80, 53.12] & 0.493 \\
\bottomrule
\end{tabular}
\end{table}

\begin{figure}[htbp]
\centering
\includegraphics[width=\linewidth]{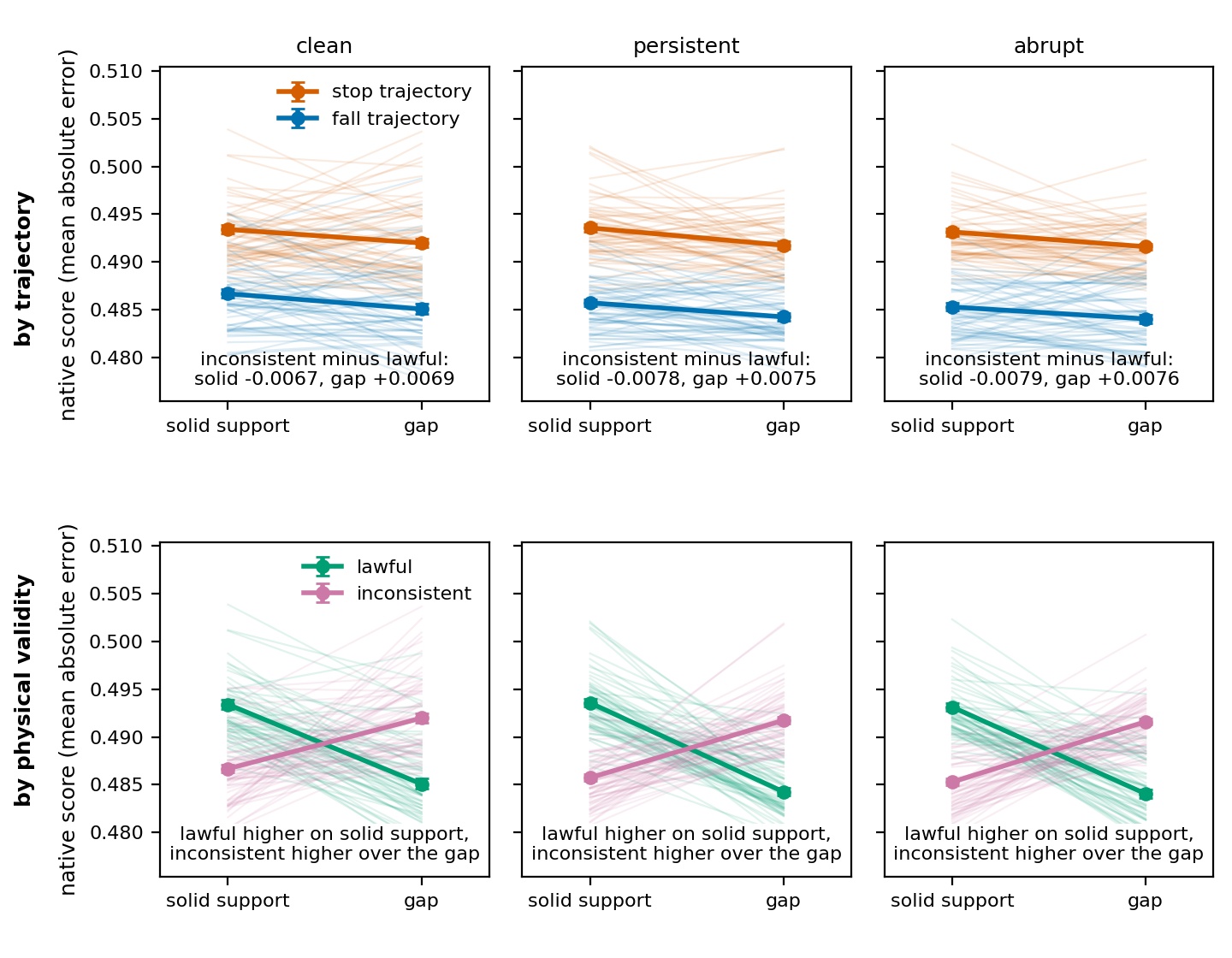}
\caption{The same final256 scores grouped by trajectory (top) and physical validity (bottom). Thick lines show means over all 256 configurations; faint lines show a fixed random display subset of 60 configurations (seed 1). Stopping has higher mean error than falling in both worlds, while the lawful/inconsistent ordering reverses when the supporting surface changes. Error bars are descriptive 95\% configuration-bootstrap intervals from 10,000 draws (seed 20261031), not the multiplicity-adjusted primary margin intervals reported above.}
\label{fig:appendix-final256-crossover}
\end{figure}

\begin{figure}[htbp]
\centering
\includegraphics[width=\linewidth]{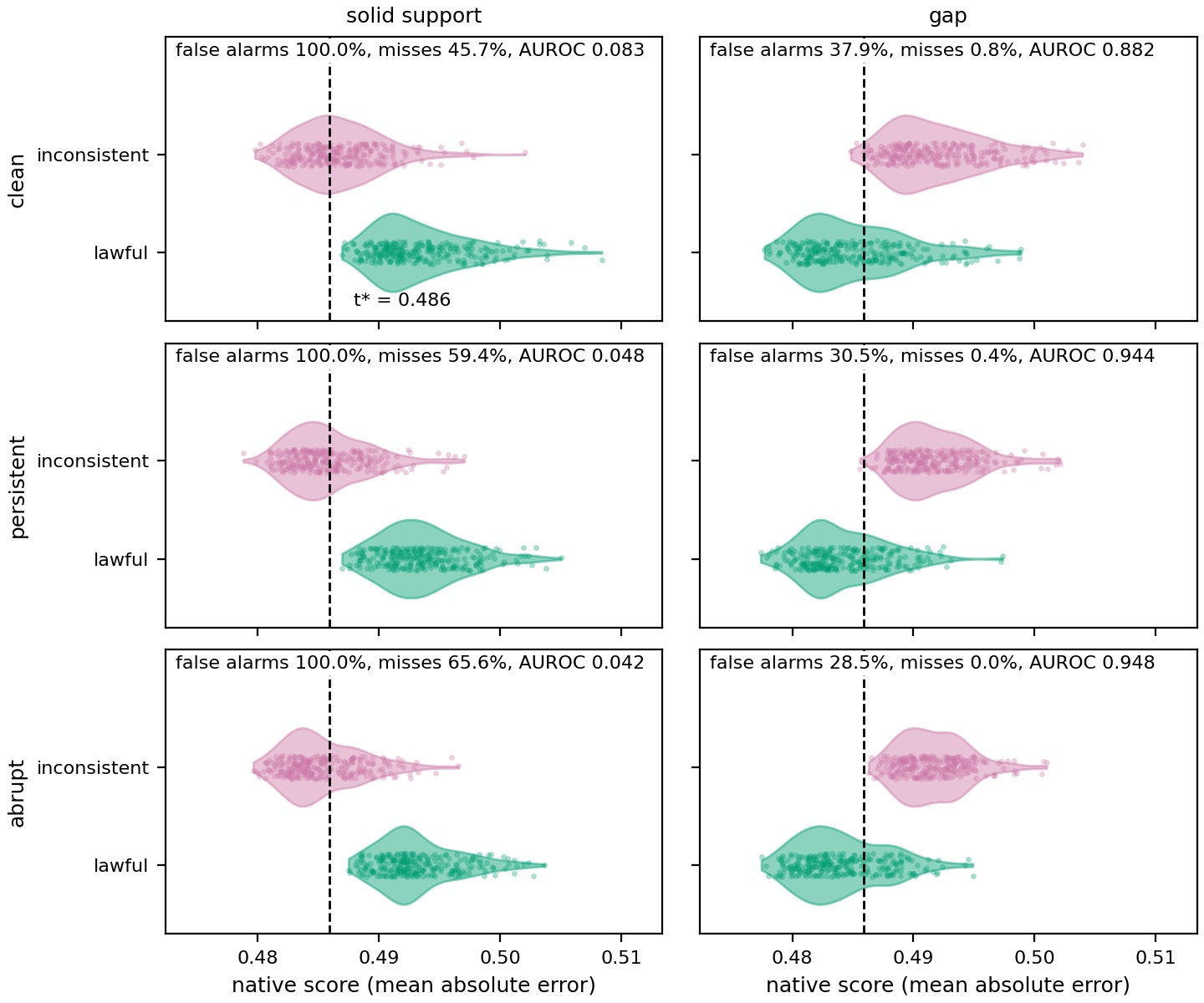}
\caption{Final256 score distributions under the unchanged development threshold (dashed line). Each class in each panel contains one score from every configuration; vertical jitter is for display only. Violins are smoothed density estimates, not confidence intervals. Scores at or above the threshold are flagged as inconsistent. The false-alarm rates, miss rates and AUROC match Table~\ref{tab:final256-thresholds}: the threshold flags every lawful solid-support stop, whereas performance is substantially better over the gap.}
\label{fig:appendix-final256-distributions}
\end{figure}

\FloatBarrier
\subsection{Kubric exact rerendering}
\label{app:kubric-details}
\begin{figure}[htbp]
\centering
\includegraphics[width=0.9\linewidth]{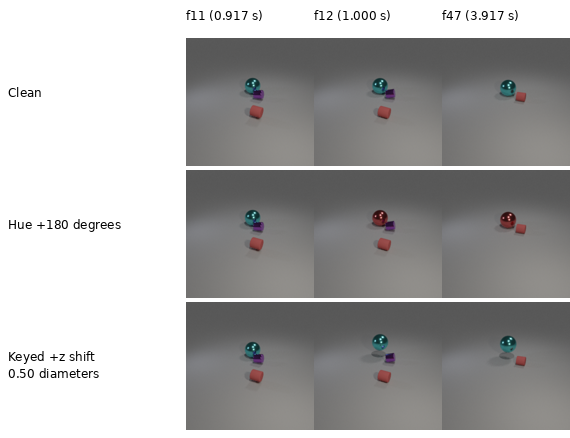}
\caption{Actual rendered inputs from the historical eight-scene Kubric pilot (seed 230207). Columns show frames 11, 12, and 47 at 12 fps. Rows show the clean render, a $180^\circ$ target-color change, and a persistent keyed upward shift of 0.50 object diameters beginning at frame 12. The first 12 frames share clean context. These archived review frames illustrate the earlier direct-displacement study, not the crossed-support experiment.}
\label{fig:kubric-pilot-examples}
\end{figure}
The Kubric pilot uses V-JEPA2-G with eight accepted MOVi-A scenes rendered from saved simulator trajectories (Figure~\ref{fig:kubric-pilot-examples}). The first 12 frames are identical context bytes across all conditions, and the target is the remaining 36 frames represented as 18 two-frame tubelets. Appearance conditions rotate only the target object's post-event base color; physical conditions apply a persistent target-object $+z$ displacement of 0.25, 0.50, 1.00, or 1.50 object diameters while preserving non-target states. The scalar score is mean absolute error between predictor output and layer-normalized EMA-target-encoder tokens. We report scene-level paired differences from clean with 10,000 bootstrap draws; with eight scenes, the matched hue-versus-displacement contrasts are exploratory and used only to test scalar specificity under exact state and rendering controls. The two predeclared physical minus appearance contrasts are $0.000259$ [$-0.000632$, $0.001145$] for a $120^\circ$ hue change versus a displacement of 0.25 object diameters, and $0.000201$ [$-0.000791$, $0.001035$] for a $180^\circ$ hue change versus a displacement of 0.50 object diameters. Both intervals include zero.

\FloatBarrier

\section{Geometry-to-error regression}
\subsection{Development-only geometry fit, evaluated post-hoc}
\label{app:geometry-transfer}
We extract FP32 target-token means from all 3,072 original final inputs at \texttt{block\_09}, \texttt{block\_19}, \texttt{block\_29}, final encoder, and native target: a $(3072,5,1408)$ matrix.
Two tables retain six physical-margin cells or eight appearance-change cells per configuration. The physical baseline uses gap, persistent, and abrupt indicators; the appearance baseline uses gap, falling-trajectory, and abrupt indicators. Both add differences in static, motion, and shuffled-frame statistics. Geometry-only models retain the same three indicators and add relative feature distance $\|z_a-z_b\|/\|z_b\|$ and its gap (physical) or abrupt (appearance) interaction. Combined models add those two terms to the full baseline. All models include an intercept and use development-only ordinary least squares (\texttt{lstsq}, \texttt{rcond=None}), with no selected regularization or layer. Coefficients were frozen before final extraction, but after final outcomes had been inspected: this is post-hoc, not an untouched confirmation.

We report MSE and $R^2=1-\operatorname{MSE}(\hat y,y)/\operatorname{MSE}(\bar y_{\rm dev},y)$, which uses the development training mean rather than the final mean. Negative values are retained. All 50,000 configuration-bootstrap draws (seed 2026102801) preserve conditions and layers. Coefficients stay fixed, so intervals omit development-fit uncertainty. The ten incremental-$R^2$ contrasts use 99.5\% intervals (Bonferroni family of ten); zero-crossing intervals do not demonstrate equivalence.

\begin{figure}[htbp]
\centering
\includegraphics[width=\linewidth]{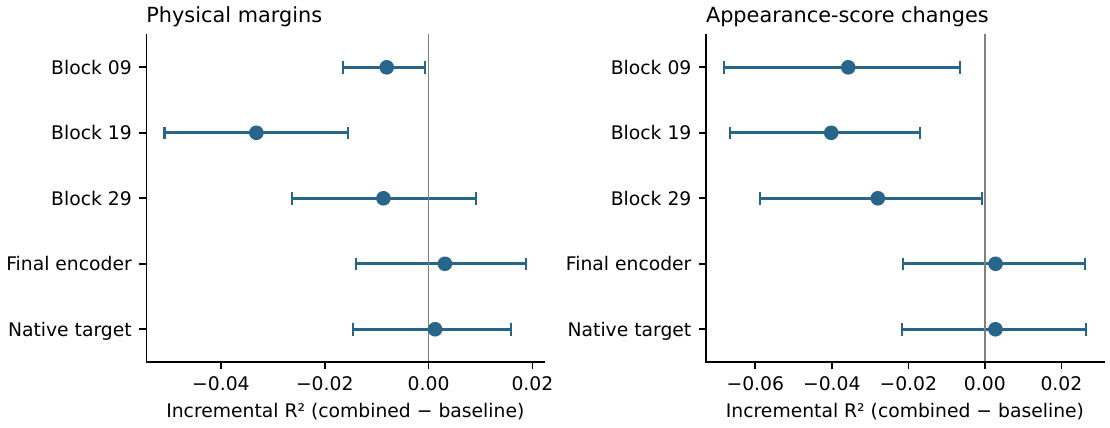}
\caption{All five fixed geometry locations and both tasks (post-hoc analysis). Bars give 99.5\% configuration-bootstrap intervals for combined-minus-baseline $R^2$, conditional on development coefficients. No positive lower bound supports incremental prediction.}
\label{fig:geometry-transfer}
\end{figure}
\begin{table}[htbp]
\centering
\footnotesize
\setlength{\tabcolsep}{3pt}
\caption{Physical margins: all fixed locations on the 256 final configurations. Baseline $R^2=0.7713$. Incremental intervals are 99.5\% configuration-bootstrap intervals. Out-of-range counts concern any combined-design value outside its development range, out of 1536 rows; no row is excluded.}
\label{tab:geometry-physical}
\begin{tabular}{@{}lrrcr@{}}
\toprule
Location & Geometry-only $R^2$ & Combined $R^2$ & Incremental $R^2$ [interval] & Out of range \\
\midrule
Block 09 & 0.7580 & 0.7632 & $-0.0081$ [$-0.0165$, $-0.0006$] & 67 \\
Block 19 & 0.7511 & 0.7381 & $-0.0332$ [$-0.0509$, $-0.0155$] & 208 \\
Block 29 & 0.7642 & 0.7626 & $-0.0087$ [$-0.0264$, $+0.0092$] & 137 \\
Final encoder & 0.7671 & 0.7744 & $+0.0031$ [$-0.0140$, $+0.0188$] & 57 \\
Native target & 0.7653 & 0.7725 & $+0.0012$ [$-0.0146$, $+0.0159$] & 57 \\
\bottomrule
\end{tabular}
\end{table}
\begin{table}[htbp]
\centering
\footnotesize
\setlength{\tabcolsep}{3pt}
\caption{Appearance-score changes: all fixed locations on the 256 final configurations. Baseline $R^2=-0.0090$. Incremental intervals are 99.5\% configuration-bootstrap intervals. Out-of-range counts concern any combined-design value outside its development range, out of 2048 rows; no row is excluded.}
\label{tab:geometry-appearance}
\begin{tabular}{@{}lrrcr@{}}
\toprule
Location & Geometry-only $R^2$ & Combined $R^2$ & Incremental $R^2$ [interval] & Out of range \\
\midrule
Block 09 & -0.0175 & -0.0447 & $-0.0357$ [$-0.0682$, $-0.0065$] & 81 \\
Block 19 & -0.0491 & -0.0492 & $-0.0402$ [$-0.0667$, $-0.0171$] & 56 \\
Block 29 & -0.0407 & -0.0370 & $-0.0280$ [$-0.0588$, $-0.0007$] & 75 \\
Final encoder & -0.0119 & -0.0062 & $+0.0028$ [$-0.0214$, $+0.0263$] & 46 \\
Native target & -0.0119 & -0.0062 & $+0.0028$ [$-0.0215$, $+0.0263$] & 47 \\
\bottomrule
\end{tabular}
\end{table}
\begin{table}[htbp]
\centering
\small
\setlength{\tabcolsep}{3pt}
\caption{Shuffled-geometry combined-model $R^2$: mean and central 95\% range of 1,000 whole-configuration permutations. These are descriptive permutation distributions, not confidence intervals or causal tests.}
\label{tab:geometry-shuffle}
\begin{tabular}{@{}lcc@{}}
\toprule
Location & Physical margins & Appearance-score changes \\
\midrule
Block 09 & 0.7671 [0.7623, 0.7725] & -0.0751 [-0.0997, -0.0492] \\
Block 19 & 0.7185 [0.7072, 0.7304] & -0.0293 [-0.0452, -0.0128] \\
Block 29 & 0.7231 [0.7098, 0.7364] & -0.0341 [-0.0535, -0.0141] \\
Final encoder & 0.7518 [0.7442, 0.7603] & -0.0316 [-0.0471, -0.0159] \\
Native target & 0.7540 [0.7469, 0.7618] & -0.0315 [-0.0470, -0.0157] \\
\bottomrule
\end{tabular}
\end{table}
\FloatBarrier

The shuffle diagnostic uses 1,000 whole-configuration permutations (seed 2026102803+$j$), identical across conditions, locations, and tasks, retaining baseline features. Its combined-model $R^2$ distribution is descriptive, not a causal or confirmatory randomization test. Extrapolation counts flag rows with any design value outside its development range; rows are not discarded or refitted. No result demonstrates that removing appearance improves native error or that the model lacks physical information.

\section{Reproducibility resources}
\label{app:static-calibration}
\label{app:resources-provenance}

{\raggedright
Code and numeric experiment records are available at
\url{https://github.com/3-ml/jepa-surprise/tree/local-launchers-2026-10-31}.\par}

\begin{itemize}
\raggedright
\item Representation and action studies: \href{https://github.com/3-ml/jepa-surprise/tree/local-launchers-2026-10-31/historical/results/}{\nolinkurl{historical/results/}}.
\item Crossed-support experiment: \href{https://github.com/3-ml/jepa-surprise/tree/local-launchers-2026-10-31/release/final256/}{\nolinkurl{release/final256/}}.
\item Physion++ controls: \href{https://github.com/3-ml/jepa-surprise/tree/local-launchers-2026-10-31/release/physion-static/}{\nolinkurl{release/physion-static/}} and \href{https://github.com/3-ml/jepa-surprise/tree/local-launchers-2026-10-31/release/physion-appearance/}{\nolinkurl{release/physion-appearance/}}.
\item Geometry-to-error regression: \href{https://github.com/3-ml/jepa-surprise/tree/local-launchers-2026-10-31/release/geometry-transfer/}{\nolinkurl{release/geometry-transfer/}}.
\item Hardware, software, and recorded runtimes: \href{https://github.com/3-ml/jepa-surprise/blob/local-launchers-2026-10-31/docs/reproducibility.pdf}{\nolinkurl{docs/reproducibility.pdf}}.
\item Checkpoint identities: \href{https://github.com/3-ml/jepa-surprise/blob/local-launchers-2026-10-31/docs/reproducibility.md#checkpoint-and-model-identities}{\nolinkurl{docs/reproducibility.md}}.
\item Execution instructions and required inputs: \href{https://github.com/3-ml/jepa-surprise/blob/local-launchers-2026-10-31/docs/LOCAL_EXECUTION.md}{\nolinkurl{docs/LOCAL_EXECUTION.md}} and \href{https://github.com/3-ml/jepa-surprise/blob/local-launchers-2026-10-31/release/DEPENDENCIES.md}{\nolinkurl{release/DEPENDENCIES.md}}.
\item File hashes and verification: \href{https://github.com/3-ml/jepa-surprise/blob/local-launchers-2026-10-31/release-manifest.json}{\nolinkurl{release-manifest.json}} and \href{https://github.com/3-ml/jepa-surprise/blob/local-launchers-2026-10-31/tools/check_release.py}{\nolinkurl{tools/check_release.py}}.
\end{itemize}

Datasets, model weights, rendered videos, and large feature caches must be obtained separately. The execution instructions specify the required input files, software versions, and hardware.
\FloatBarrier

\end{document}